\documentclass{article}

\usepackage{microtype}
\usepackage{graphicx}
\usepackage{subfigure}
\usepackage{booktabs} % for professional tables
\makeatletter
\renewcommand \thesection{S\@arabic\c@section}
\makeatother
\newcommand{\beginsupplement}{%
        \setcounter{table}{0}
        \renewcommand{\thetable}{S\arabic{table}}%
        \setcounter{figure}{0}
        \renewcommand{\thefigure}{S\arabic{figure}}%
        \setcounter{section}{0}
        \renewcommand{\thesection}{S\arabic{section}}
     }
\usepackage{hyperref}

\usepackage[accepted]{icml2021}

\icmltitlerunning{}

\begin{document}

\twocolumn[
\icmltitle{Clinical outcome prediction under hypothetical interventions – a representation learning framework for counterfactual reasoning}



\icmlsetsymbol{equal}{*}

\begin{icmlauthorlist}
\icmlauthor{Yikuan Li}{to,go}
\icmlauthor{Mohammad Mamouei}{to,go}
\icmlauthor{Shishir Rao}{to,go}
\icmlauthor{Abdelaali Hassaine}{to,go}
\icmlauthor{Dexter Canoy}{to,go,ed}
\icmlauthor{Thomas Lukasiewicz}{pd}
\icmlauthor{Kazem Rahimi}{ed,to,go}
\icmlauthor{Gholamreza Salimi-Khorshidi}{to,go}
\end{icmlauthorlist}

\icmlaffiliation{to}{Deep Medicine, Oxford Martin School, University of Oxford, Oxford, United Kingdom}
\icmlaffiliation{go}{Nuffield Department of Women’s \& Reproductive Health, University of Oxford, Oxford, United Kingdom}
\icmlaffiliation{ed}{NIHR Oxford Biomedical Research Centre, Oxford University Hospitals NHS Foundation Trust, Oxford, United Kingdom}
\icmlaffiliation{pd}{Department of Computer Science, University of Oxford, Oxford, United Kingdom}

\icmlcorrespondingauthor{Yikuan Li}{yikuan.li@wrh.ox.ac.uk}
% \icmlcorrespondingauthor{Eee Pppp}{ep@eden.co.uk}

\icmlkeywords{Electronic health records, risk prediction, counterfactual reasoning}

\vskip 0.3in
]



\printAffiliationsAndNotice{}

\begin{abstract}
Most machine learning (ML) models are developed for prediction only; offering no option for causal interpretation of their predictions or  parameters/properties. This can hamper the health systems’ ability to employ ML models in clinical decision-making processes, where the need and desire for predicting outcomes under hypothetical interventions (i.e., counterfactual reasoning/explanation) is high. In this research, we introduce a new representation learning framework (i.e., partial concept bottleneck), which considers the provision of counterfactual explanations as an embedded property of the risk model. Despite architectural changes necessary for jointly optimising for prediction accuracy and counterfactual reasoning, the accuracy of our approach is comparable to prediction-only models. Our results suggest that our proposed framework has the potential to help researchers and clinicians improve personalised care (e.g., by investigating the hypothetical differential effects of interventions).
\end{abstract}

\section{Introduction}
Recent advances in deep learning (DL) -- a subfield of machine learning (ML) -- has led to great hopes for its potential impact in medicine. One of the key reasons behind DL methods' success and popularity, is their ability to {\em learn} useful representations (e.g., of patients, or images, or sentences) from raw or minimally-processed input data. In other words, instead of relying on expert-provided representations (or, feature engineering), most DL paradigms employ a large number of transformations, across multiple layers, to map their high-dimensional inputs to (much) lower-dimensional outputs~\cite{Goodfellow-et-al-2016}. Success in such mappings implies that a model has learned useful representations of its input data -- and hence DL being commonly referred to as "representation learning".

Most DL frameworks are what we can refer to as "straight-through representation learning" (STRL): They map an input $x$ to an output $y$, via representation $l$ that is not constrained (e.g., to correspond to certain {\em a priori} known concepts). That is, while entries in $l$ might happen to correspond to variables familiar to researchers/users (e.g., presence of a dog in an image, or, cardio-metabolic health issues in a patient), there is no guarantee for this to be the case. While the presence of such correspondences may not be important in many prediction-only tasks, in many high-stake decision making (such as those in healthcare), the one's inability to understand and interact with such complex representations (e.g., through comparing them against the external knowledge and observation) is essential. Therefore, any solution that can help provide simple insights about such DL models, can help domain experts better understand the rationale behind the predictions and act with better insights.

Suppose that a doctor is using a DL model to predict the risk of an event such as heart failure (HF). It is common for such a doctor to expect some form of explanation about the model’s predictions: What factors have the biggest influence on the predictions? Are there any interventions that can reduce certain risks for some patients? Could a different set of events (such as diagnosis, medication, or intervention) in a patient's record change his/her current predictions? 

Based on Pearl and Mackenzie’s~\cite{pearl2018book} three-level “causal ladder”, one can classify such explanations under association, intervention, and counterfactual categories. Most current methods for providing insights about DL models operate at the lowest level of the causal ladder (i.e., association or prediction); they learn a joint distribution of variables, which enables them to then predict the likelihood of a given value of the output variable $y$, conditional on the other (i.e., input) variables $x$~\cite{molnar2019,yoon2018invase,li-etal-2016-visualizing}.  The second level of the causal ladder (i.e., intervention) introduces an additional level of complexity, which captures the change in the joint distribution that can result from an exogenous intervention~\cite{pearl2009causality}; for instance, change in the price of an item can lead to change in its customers' behaviour, which can make the historical sales data less relevant for modelling future sales. Note that, given that the second-level distribution under no intervention is equal to the original distribution, the intervention level subsumes the prediction level in the ladder. While the second level of the ladder allows us to condition on variables after intervention, the third level allows us to condition on variable before intervention, by allowing us to ask “counterfactual” questions~\cite{lewis2013counterfactuals}. At this level, we can ask questions such as: How much CVD (cardiovascular disease) mortality would have been delayed, if anti-hypertensive medications were prescribed one year earlier than the date they were prescribed? That is, instead of the “factual” observations, we are interested in an alternative reality (or a counterfactual scenario), which was not actually observed. 

There are recent developments in ML that investigate the explanations at the second and third levels of the ladder -- answering the “do” and “what if” questions by conducting post-hoc interventions as a separate tool on high-level (latent or human understandable) concepts~\cite{goyal2019explaining,BAM2019,NEURIPS2020_3a93a609}, the nature of counterfactual questions, however, makes a model-free approach implausible. Therefore, in a recent study, Koh et al.~\cite{koh2020concept} proposed a framework called concept bottleneck model, which considers an intervention pathway as an inherent component of risk prediction model. In this approach, instead of the common STRL approaches, the model first learns to predict “concepts” (i.e., intermediate labels provided by humans) from inputs such as images (i.e., referred to as bottleneck); next, these learned concepts are mapped to the outputs/predictions. This enables a correspondence between latent representations and concepts, which then allows one to conduct interventions on concepts/representations (e.g., what if we change a certain concept’s value from 0 to 1?) to investigate their effects on the prediction/outcome. One of the drawbacks of this framework is that a complete list of pre-defined concepts is needed to ensure sufficient predictive power (i.e., accuracy); the more comprehensive the pre-defined concepts, the better the prediction. %In other words, assume that there is a (lower dimensional) concept vector that can be optimally derived from the (higher dimensional) input vector, through some transformations; losing any of the entries/concepts in this concept vector (even the ones that are not of interest for conducting an intervention), will imply the loss of input information, and hence a drop in prediction power/accuracy. 

In counterfactual analyses, the degree to which one can change some aspects of the input should be consistent with the empirical evidence (i.e., “plausible counterfactuals”)~\cite{pearl2009causality}. For instance, we will be less likely to simply add heart failure to a young patient’s counterfactual record, if the empirical evidence suggests that HF almost always happens at older ages; similarly, if the empirical evidence suggests that diabetes is very likely to happen in a particular group of patients with certain characteristics, it is more plausible to add diabetes to such patients’ counterfactual records. The assessment of these plausibility criteria can have varying degree of difficulty, when going from one domain to another. In epidemiology and causal inference, the commonly used methods to ensure the plausibility of counterfactuals are randomisation and propensity score matching~\cite{szklo2014epidemiology}. Both these methods must be applied for cohort selection before causal reasoning and restrict our attention to a well-specified causal question (e.g., how a concept affects an outcome?). This makes them less suitable for conducting counterfactual explanations for a risk prediction model that works for the general population with multiple interventions of interest. Therefore, a risk model that can provide accurate risk prediction for the general population and with the ability to provide counterfactual explanations for individuals seems to be the most ideal solution for guiding clinical decision-making.

\begin{figure}[h!]
\vskip -0.1in
\begin{center}
\centerline{\includegraphics[width=\columnwidth]{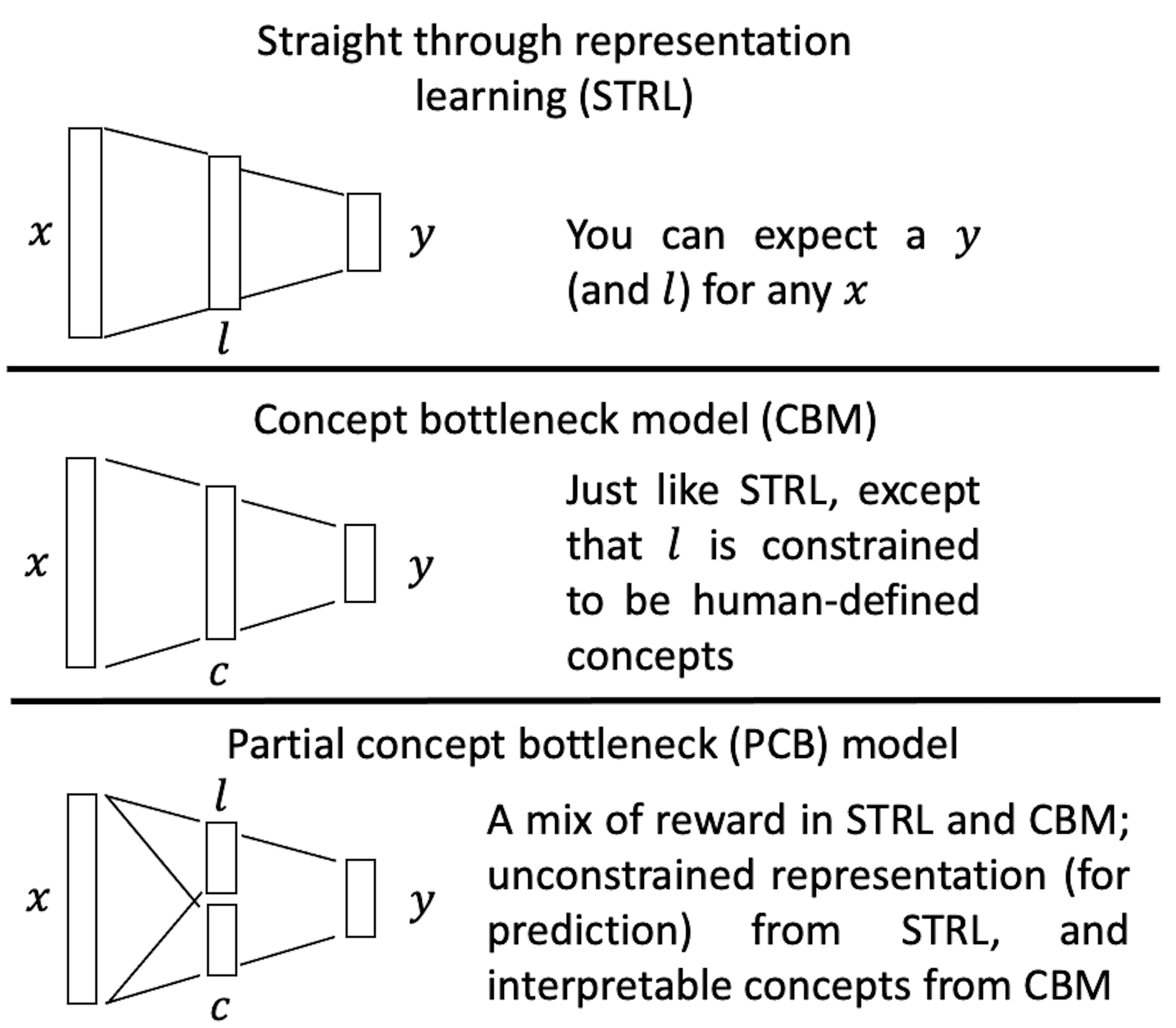}}
\vskip -0.1in
\caption{STRL, concept bottleneck, and PCB models. The representations learned in STRL models are usually not human-understandable; in contrast, concept bottleneck model first maps the raw features to human-understandable/-defined concepts $c$, then it predicts output using $c$. While the concept layer greatly improves model interpretability, the comprehensiveness of $c$ is essential to retain accurate prediction. The PCB framework predicts an intermediate set of human-specified concepts of interest, $c$, plus a set of discrete latent features, $l$; then it uses both $c$ and $l$ to predict the final output $y$. Compared to STRL and concept bottlenck model, PCB adds a constraint to have the human-defined concepts as part of the intermediary representations, which enable further investigations such as counterfactual reasoning. Furthermore, the latent representation used by PCB capture features that can preserve the predictive accuracy.}
\label{fig:central_illustration}
\end{center}
\vskip -0.2in
\end{figure}

In this study, we aim to build on recent developments in concept bottleneck models and propose a new framework of partial concept bottleneck (PCB). Our results show that PCB provides accurate predictions (i.e., comparable to the black-box models), while enabling counterfactual reasoning under hypothetical interventions. We aim to employ this PCB framework for risk prediction using electronic health records (EHRs). PCB (shown in Figure~\ref{fig:central_illustration}) allows us to incorporate pre-defined concepts (e.g., potential interventions, derived from expert knowledge) as an inherent component in risk prediction model. Given such an architecture, PCB can take advantage of DL to learn representations that can (1) preserve the alternative STRL models' predictive power, and (2) phenotype patients into clusters within which patients can be investigated for plausible counterfactuals. We tested this PCB framework in two counterfactual reasoning experiments; while providing SOTA-level predictive performance, we demonstrated how our PCB framework can provide simple insights about its inner workings through counterfactual explanations.

\section{Materials and Methods}
\subsection{Dataset}
In this study, we used the data from Clinical Practice Research Datalink (CPRD)~\cite{herrett2015data}. CPRD is one of the largest de-identified linked longitudinal EHR datasets collected from general practices (GPs) across the UK. CPRD contains data from approximately 7\% of the UK population~\cite{conrad2018temporal}, and can be linked to other secondary care datasets such as Hospital Episode Statistics (HES), Admitted Patient Care data, and data from the Office for National Statistics. In order to select the appropriate cohort for this study, in brief, we included patients who: (1) were 16 or older and contributed to data between Jan 1, 1985, and Sep 30, 2015; (2) were eligible to be linked to HES; and (3) had “accept” status from CPRD quality control~\cite{herrett2015data}. Records collected from diagnoses, medications, GP tests, hospital procedures, blood pressure measurement, body mass index, smoking status, and drinking status were used for modelling. More on cohort selection can be found in Supplementary, Method S1.

\subsection{Concept bottleneck model}
Unlike the STRL models that predict the output $y$ directly from the input $x$ (i.e., $x$→$y$), the concept bottleneck model~\cite{koh2020concept} constructs an intermediate layer between the $x$ and $y$ with only high-level concepts $c$ (i.e., $x$→$c$→$y$). The concept bottleneck model learns to both predict the concepts from the input, and to use the concepts to predict the outputs. Since the intermediate layer is expected to capture all the useful information in $x$ in the form of high-level concepts (i.e., through forming an information bottleneck), it is also called a bottleneck layer. Mapping inputs to the concepts, requires human-generated annotations for the concepts (somewhat similar to the classic feature engineering). The mapping from concepts to the outputs can be done using a classifier (or regressor). Therefore, in theory, one can manipulate the value of concepts and observe its effect on the prediction and hence provide model interpretations and counterfactual explanations. 

\subsection{Discrete representation learning}
Representation learning, especially for learning continuous representations, enables DL models to extract latent features that can be useful for various tasks. One of the challenges of dealing with such representations is the lack of a priori known correspondence between their values and real-world concepts. That is, their continuous range, can hamper one’s ability to find simple meaning and interpretation in their values. Learning discrete representations, on the other hand, can provide a simpler solution for reasoning and scenario analysis in the representation space (e.g., for model interpretation). 

Additionally, in such discrete representation spaces, the number of patient clusters will be smaller (compared to the continuous space), and more interpretable. This allows one to reason about hypothetical interventions in each cluster to further investigate various what-if questions. The main concern regarding the use of discrete (instead of continuous) representations is the loss of information in the mapping and hence less accurate risk estimation. Therefore, it is important to investigate the trade-off between the gain in interpretability and reasoning, against the potential loss in model accuracy. 

\begin{figure}[ht]
\vskip -0.1in
\begin{center}
\centerline{\includegraphics[width=0.75\columnwidth]{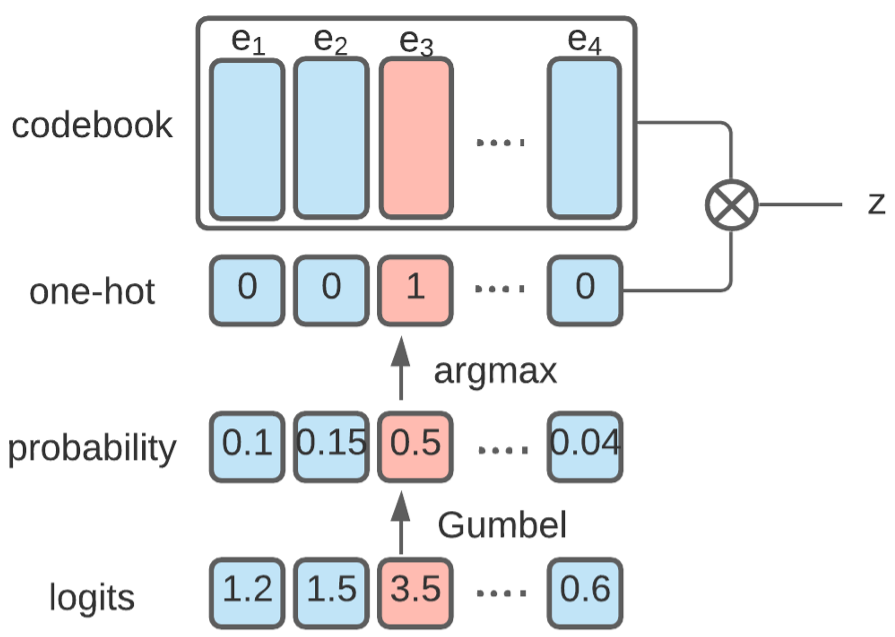}}
\vskip -0.1in
\caption{Vector quantisation for a categorical variable. The logits are transformed into a one-hot vector using Gumbel-Softmax quantisation and represented with corresponding embedding representation in the codebook.}
\label{fig:VQ}
\end{center}
\vskip -0.3in
\end{figure}

An effective approach for learning discrete representations is vector quantisation~\cite{Baevski2020vq-wav2vec:,NIPS2017_7a98af17}. It uses differentiable Gumbel-Softmax distribution to approximate the Bernoulli or categorical distribution; it then uses a codebook to map the discrete representation to the embedding space. Figure~\ref{fig:VQ} shows discrete representation learning, where logits are transformed into one-hot vectors and represented with corresponding embedding representation (z) in the codebook. A more practical solution, however, is to separate the logits into n groups, then apply the vector quantisation to each group. Therefore, the logits can be transformed into n binary or categorical variables, and an n-dimensional binary representation can maximumly cluster observations (i.e., patients) into $2^n$ groups. In this work, n can be tuned as a hyperparameter (based on model performance). Ideally, we would like to have n as small as possible, without substantially sacrificing the model accuracy (i.e., the bias-variance trade-off). 

\subsection{Partial concept bottleneck model}
Concept bottleneck models require a comprehensive list of concepts for accurate prediction. However, the definition and selection of a complete list of concepts are difficult. It not only requires a significant amount of expert labour, but also highly relies on how the underlying causes of a condition are understood today. Even with a complete list of concepts, due to substantial data missingness (random or not at random) in EHR, the ground-truth concept labels are not always available (e.g., not available in a dataset, or missing for a large proportion of the population). These restrictions hinder the usability of concept bottleneck models in EHR applications.

In this work, we propose PCB (as shown in Figure~\ref{fig:central_illustration}) as a solution; it substantially relaxes the restrictions that one faces when using concept-based models for counterfactual reasoning. It modifies concept bottleneck models of representation learning, by introducing a combination of $l$ , and concept representation $c$, which corresponds to human-defined concepts of interest (for hypothetical interventions). Compared to the simple concept bottleneck models, which require a large amount of work on concept annotation and a robust list of concepts to preserve predictive performance, $l$ takes care of the rest of the latent feature extraction. 

The PCB model can be trained in a similar way to the concept bottleneck models. It includes three components, a function $g(x)$ that maps $x$ to $c$, a function $h(x)$ that maps $x$ to $l$, and a classifier $f(c,l)$ that predicts $y$ based on $c$ and $l$. This results in the following optimisation objective:\(\hat{f},\hat{g},\hat{h}=\arg\min L_{Y}(f(c,h(x)),y)+L_{C}(g(x),c)\). While $\hat{f}$ is trained with the ground truth concepts, it still takes the $\hat{g}(x)$ as input at test time. Additionally, the objective can be modified as below if having a model $m(\cdot)$ for representation learning before mapping to $c$ and $l$: \(\hat{f},\hat{g},\hat{m},\hat{h}=\arg\min L_{Y}(f(c,h(m(x))),y)+L_{C}(g(m(x)),c)\).

% i am writing equation in this way to save the space, cuz we exceed the page limitation. We do not refer this equation in the paper, so i think should be fine ? 

\subsection{Baseline models}
In this study, we compared the performance of PCB with standard DL models for HF risk prediction, to investigate the trade-off between counterfactual reasoning and prediction accuracy. For PCB models, we constructed the initial representation learning architecture $m(\cdot)$ by adopting a previous high-performing Transformer~\cite{DBLP} model architecture, Hi-BEHRT, and its parameters~\cite{li2021hi} (see Supplementary, Method S2 for more). Next, a two-layer multi-layer perceptron $g(\cdot)$ and a vector quantization component $h(\cdot)$ were trained to map $m(x)$ to $c$ and $l$, respectively. We used vector quantization that maps representations into n binary variables for all tasks. The classifier $f(\cdot)$ that predicts $y$ based on $c$ and $l$ is trained with a three-layer multi-layer perception. Additionally, Hi-BEHRT is used as a standard end-to-end black-box model for comparison for each task. See supplementary, Method S3 for more details on the experimental settings.

\subsection{Counterfactual explanations}
The PCB model establishes a relationship from $x$ to $y$ through $c$ and $l$ (i.e., $x$→$c$, $l$→$y$), where the path for $c$ is not contaminated with other variables. For a specific $l$, $y$ only depends on $c$. Therefore, any modification (referred to as $do(\cdot)$) on $c$ is a counterfactual from the model’s perspective ($p(y|do(c),l)$). $c$ does not have to be a factor that causes $y$ in the narrower biologically or medically defined causal web. For example, the frequency of hospital admission does not cause HF; nevertheless, it can be predictive and is associated with HF. Therefore, we can investigate the counterfactuals in terms of different frequencies of hospital admission by hypothetically changing this concept.

\begin{figure}[h!]
\vskip -0.1in
\begin{center}
\centerline{\includegraphics[width=0.9\columnwidth]{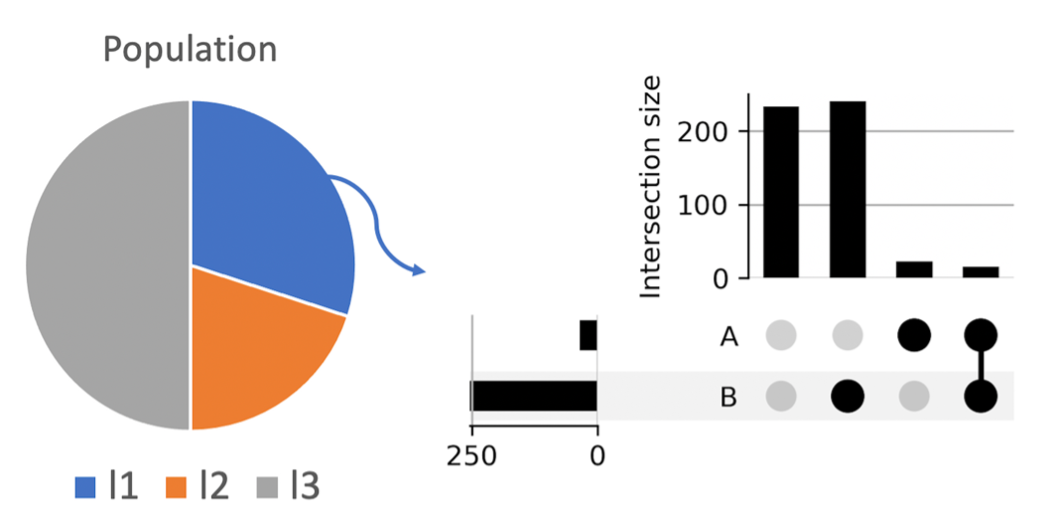}}
\vskip -0.1in
\caption{Illustration of UpSet plot for analysing counterfactual plausibility. In this toy example, patients are clustered into three groups (i.e., $l_{1}$,$l_{2}$, and $l_{3}$) by a PCB model with two concepts of interest (A and B) for a hypothetical population (left). For patients in the group $l_{1}$, the UpSet plot shows (right)that 250 and approximately ten patients have B (+) and A (+), respectively, on the x-axis, while the numbers for A(-) and B(-) are unknown.  Most of the patients within this group are A (-) and B (-) (approximately 220 patients), and A (-) and B (+) (about 225 patients) as shown on the y-axis (intersection size). The summation of the intersection size across sets is the total number of patients within the group.}
\label{fig:upset_demo}
\end{center}
\vskip -0.2in
\end{figure}

Additionally, the counterfactual questions should satisfy the counterfactual plausibility~\cite{pearl2018book}. Thus, the hypothetical intervention cannot contradict the facts (i.e., observations), and in the PCB model, the facts are restricted by $l$. For a specific cluster of patients $l_{i}$ ($i=1,…,2^n$), only plausible interventions can be conducted, and this can be represented as $p(y|do(c),l=l_{i})$. In this work, we employ the UpSet plot~\cite{lex2014upset} as the systematic solution for analysing the plausibility of counterfactuals, using the empirical evidence. As shown in Figure~\ref{fig:upset_demo}, the UpSet plot carries out a comprehensive descriptive analysis, illustrating the possibility and probability of various counterfactuals that can happen for a specific cluster of patients. For example, for a patient who is A (-) and within the cluster $l_{1}$ in Figure~\ref{fig:upset_demo}, it is plausible to ask the counterfactual question: what would be the outcome if the patient is B (+) instead of B (-)? In contrast, although it is possible to ask a question for a B (-) patient within the cluster $l_{1}$, what would be the outcome if the patient is A (+) instead of A (-)? Due to the rare prevalence in the alternative scenarios, we might consider the counterfactual as less plausible. Therefore, when asking a counterfactual question, we can consider a counterfactual is plausible if its prevalence in the group is within a certain range (i.e., threshold). The range can vary depending on the purpose of the task and what is the question at hand. 100\% and 0\% prevalence of a concept or concepts combination in a cluster means that the counterfactual is impossible for the specific cluster, hence, implausible.  

\subsection{Model evaluation and counterfactual analysis}
We randomly split the dataset into 60\% training set, 10\% tuning set, and 30\% validation set. We reported the model results with the best task accuracy in area under the receiver operating characteristics curve (AUORC) and area under the precision-recall curve (AUPRC) on the validation set. For counterfactual analysis, we selected groups that can cover approximately 95\% of the cohort. For each group, we applied the UpSet plot to analyse the plausibility of different counterfactuals and reported the estimated risk or risk ratio (RR) for each counterfactual. The sanity check for counterfactual analysis was conducted by comparing the consistency between the estimated risk ratio and the observed risk ratio for a counterfactual scenario in each group and can be found in Supplementary Methods S4. 

\section{Experiments}
The PCBs were trained for five-year incident HF risk prediction in two different experimental settings. For each setting, we use features before a defined baseline to predict the HF risk in future. HF is defined using a list of conditions reported in the previously established research~\cite{conrad2018temporal} (Table S1). The diagnosis code list for the extraction of each condition in the following sections was adapted from the CALIBER code repository~\cite{kuan2019chronological}. Below, we describe these two experiments.

\textbf{AF-HF}: This task investigates how having atrial fibrillation (AF) can influence the risk of HF in patients with hypertension and/or diabetes. In this experiment, the concepts are defined as AF, hypertension, and diabetes. We define a concept as positive if the corresponding condition is recorded in the primary care or hospital admission before the baseline. The task was trained and evaluated using the entire cohort.  

 \textbf{F-HF}: This task investigates how different frequencies of all-cause hospital admission and GP visits influence the HF risk on patients with coronary heart disease (CHD). The concepts of interest in this experiment are number of follow-up years at baseline after the incident CHD and averaged the frequency (per year) of all-cause hospital and GP visits after incident CHD. The incident CHD is defined as the first record of CHD recorded in the primary care or hospital admission records. Because this task focuses on patients with CHD, only a subset of patients with CHD in the cohort is used for model training and evaluation.

\section{Results}
\subsection{Descriptive analysis for HF cohort}
Between January 1985 and September 2015, 1,975,630 patients who met the criteria for inclusion were selected for HF risk prediction and interpretation. 1,995 diagnoses codes in international classification of diseases, tenth revision (ICD-10)~\cite{world2004international} level 4 and 378 medications codes in British National Formulary (BNF)~\cite{jointbritish} chapter level, 960 The Office of Population Censuses and Surveys (OPCS) Classification of Interventions procedure codes and 275 test codes in Read code~\cite{chisholm1990read} were included for modelling. The descriptive analysis of patients’ characteristics at baseline is shown in Supplementary, Table S2.

\subsection{Prediction performance evaluation}
To the best of our knowledge, there is no similar work to the proposed approach. Hence, we mainly compared the PCB to the standard black-box models. Table~\ref{tab:accuracy} shows that the PCB models achieve comparable performance to the standard end-to-end black-box models on both experiments. While the result proves the viability of PCBs in risk prediction, their strength is to further provide the counterfactual explanations. 

\begin{table}[h!]
\vskip -0.2in
\caption{Task performance with 95\% confidence interval over random seeds. Overall, PCBs achieve comparable performance as to standard end-to-end models. }
\label{tab:accuracy}
\begin{center}
\begin{small}
\begin{sc}
\begin{tabular}{c|c|c}
\toprule
 &\multicolumn{2}{ c }{AF-HF} \\
\midrule
 & Standard & PCB\\
\midrule
AUROC & 0.96(± 0.01)& 0.95(± 0.00)\\
AUPRC & 0.77(± 0.01)& 0.75(± 0.01)\\
\midrule
 & \multicolumn{2}{ c }{F-HF}\\
 \midrule
AUROC &  0.90(± 0.00)& 0.87(± 0.01)\\
AUPRC &  0.80(± 0.00) & 0.77(± 0.01) \\
\bottomrule
\end{tabular}
\end{sc}
\end{small}
\end{center}
\vskip -0.2in
\end{table}

\subsection{Counterfactual explanations}
\begin{figure}[h!]
\vskip -0.1in
\begin{center}
\centerline{\includegraphics[width=\columnwidth]{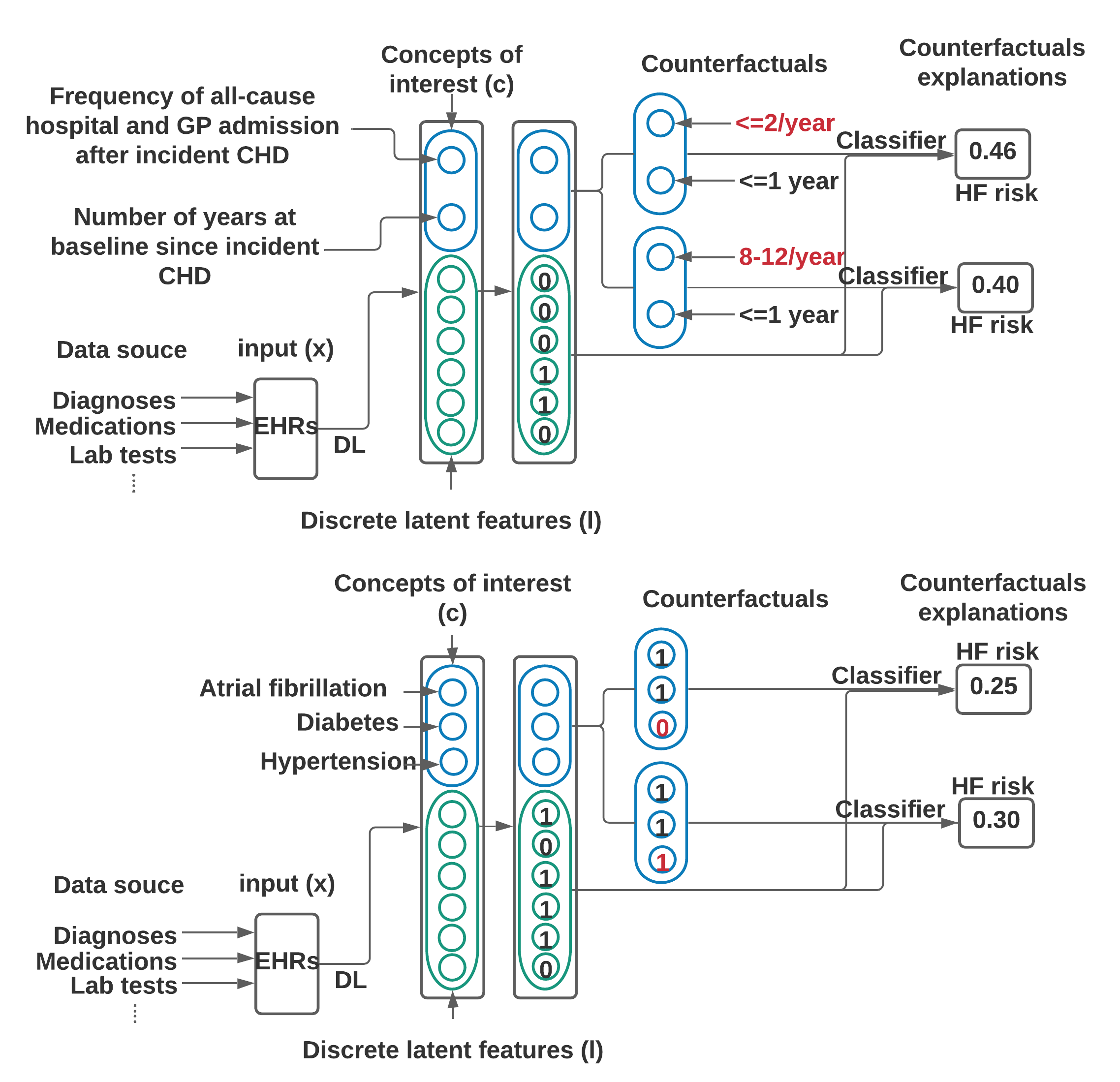}}
\vskip -0.1in
\caption{Examples of counterfactual explanations, where we intervene on the concepts of interest to investigate the “what if” question. Top (F-HF): for a group of patients with the latent feature “0,0,0,1,1,0” and with $\leq$1 year of follow-up at baseline after incident CHD, the model thinks having 8-12 visits/year would reduce the HF risk by 13\% compared to patients with $\leq$2 visits/year. Bottom (AF-HF): for patients with the latent feature “1,0,1,1,1,0”, the model thinks having hypertension as a comorbidity of atrial fibrillation and diabetes would increase the risk of HF by 20\%. }
\label{fig:counter_illu}
\end{center}
\vskip -0.3in
\end{figure}

Figure~\ref{fig:counter_illu} shows some examples of counterfactual explanations. The discrete latent representations ($l$) learned covariates that can phenotype patients into clusters. By looking at a specific group (i.e., $l_{i}$), we can intervene on the concepts of interest ($c$), by manipulating their values to inspect how the final prediction changes. The counterfactual reasoning here is: For a patient in a particular group/cluster, how would the risk of HF change if the patient had hypertension? Or, how will more frequent visit to health systems at the onset of CHD, change the reduce the HF?

\subsection{Counterfactual analysis on AF-HF}
\begin{figure}[h!]
\vskip -0.1in
\begin{center}
\centerline{\includegraphics[width=\columnwidth]{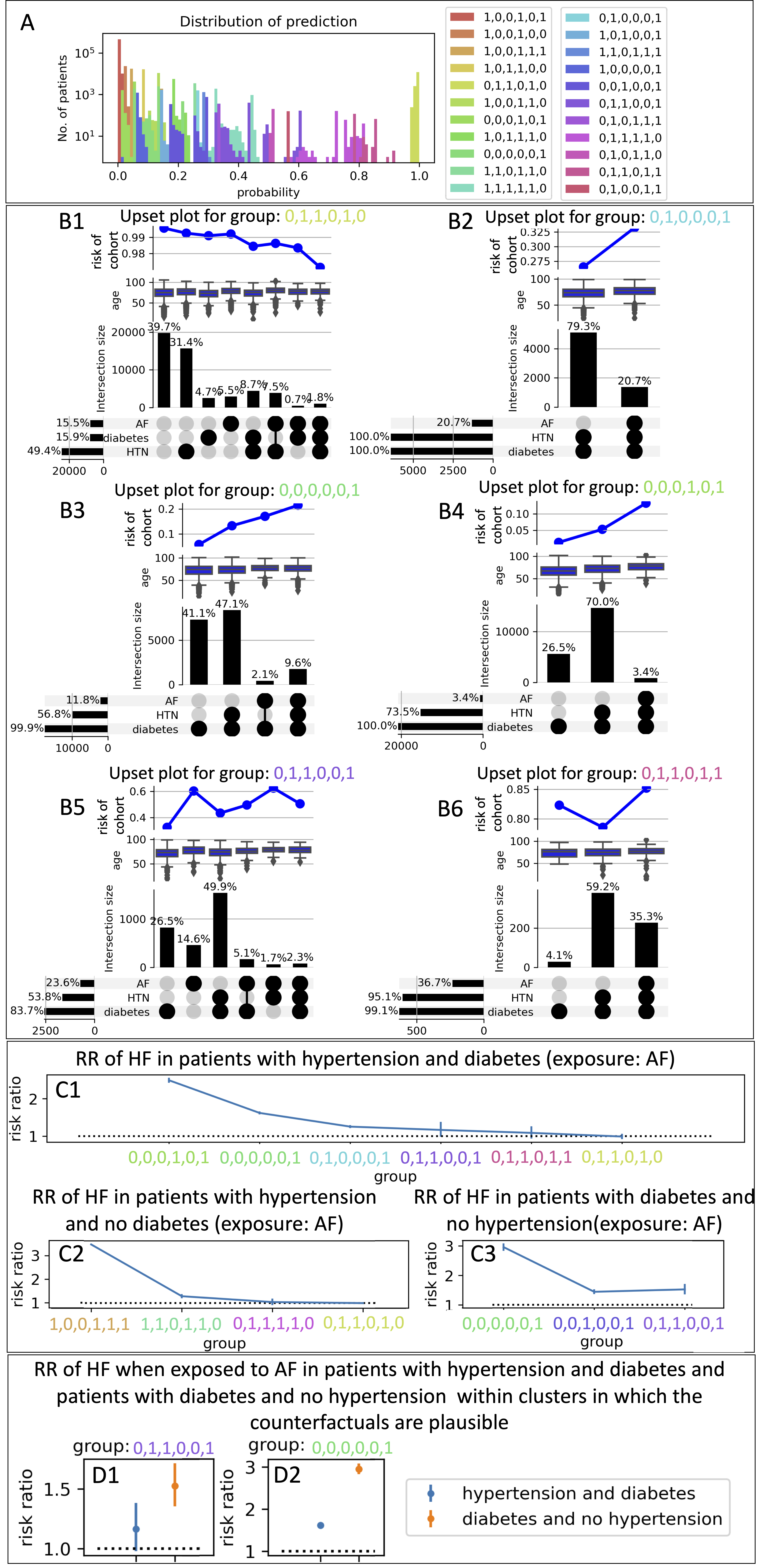}}
\vskip -0.1in
\caption{Descriptive analysis and counterfactual explanations for HF risk with respect to AF in patients with hypertension or/and diabetes. A: distribution of predicted HF risk on the validation set; the colours represent the patient clusters. B1 to B6 are UpSet plots (with age distribution and estimated HF risk) for clusters where AF is a plausible counterfactual for patients with diabetes and hypertension. C1, C2, and C3 represent the RR of HF in patients with diabetes and/or hypertension (sorted by ascending order in terms of the HF risk in each group). D1 and D2 show how AF affects HF risk on two plausible hypothetical patient groups within the same cluster.}
\label{fig:af-hf}
\end{center}
\vskip -0.3in
\end{figure}

In this experiment, we applied the PCB model to HF risk prediction on the general population with AF, diabetes, and hypertension as the concepts of interest. Figure~\ref{fig:af-hf}A shows the PCB model phenotypes the entire population into 22 clusters with a six-dimensional latent representation. Additionally, within the 22 clusters, only six (Figure~\ref{fig:af-hf} B1 to B6) have our desired counterfactuals as plausible (defined in Methods: Counterfactual explanations) and hence allow us to explicitly investigate the effect of AF on HF risk for patients with both diabetes and hypertension. The results are summarised in Figure~\ref{fig:af-hf} C1, which shows AF can increase the risk of HF across almost all clusters that have AF as a plausible counterfactual; the RR is substantially higher in clusters with lower risk of HF (e.g., “0,0,0,1,0,1” and “0,0,0,0,0,1”), and the RR decreases as the HF risk of a cluster increases (from “0,0,0,1,0,1” to “0,1,1,0,1,0”). 

Similar patterns were observed for patients with only hypertension (4 clusters) and only diabetes (3 clusters), as shown in Figure~\ref{fig:af-hf} C2 and C3. The corresponding UpSet plots can be found in the Supplementary. Additionally, D1 and D2 show a consistent pattern for patients within the same cluster that AF can increase the risk of HF more in patients with only diabetes than patients with both hypertension and diabetes. However, the absolute risk of HF is slightly higher in patients with both hypertension and diabetes. Because there is no intersection of clusters between C2 and C3, the comparison of how AF influences the HF risk in patients with only diabetes or hypertension cannot be conducted (implausible).

\subsection{Counterfactual analysis on F-HF}
\begin{figure}[h!]
% \vskip -0.2in
\begin{center}
\centerline{\includegraphics[width=\columnwidth]{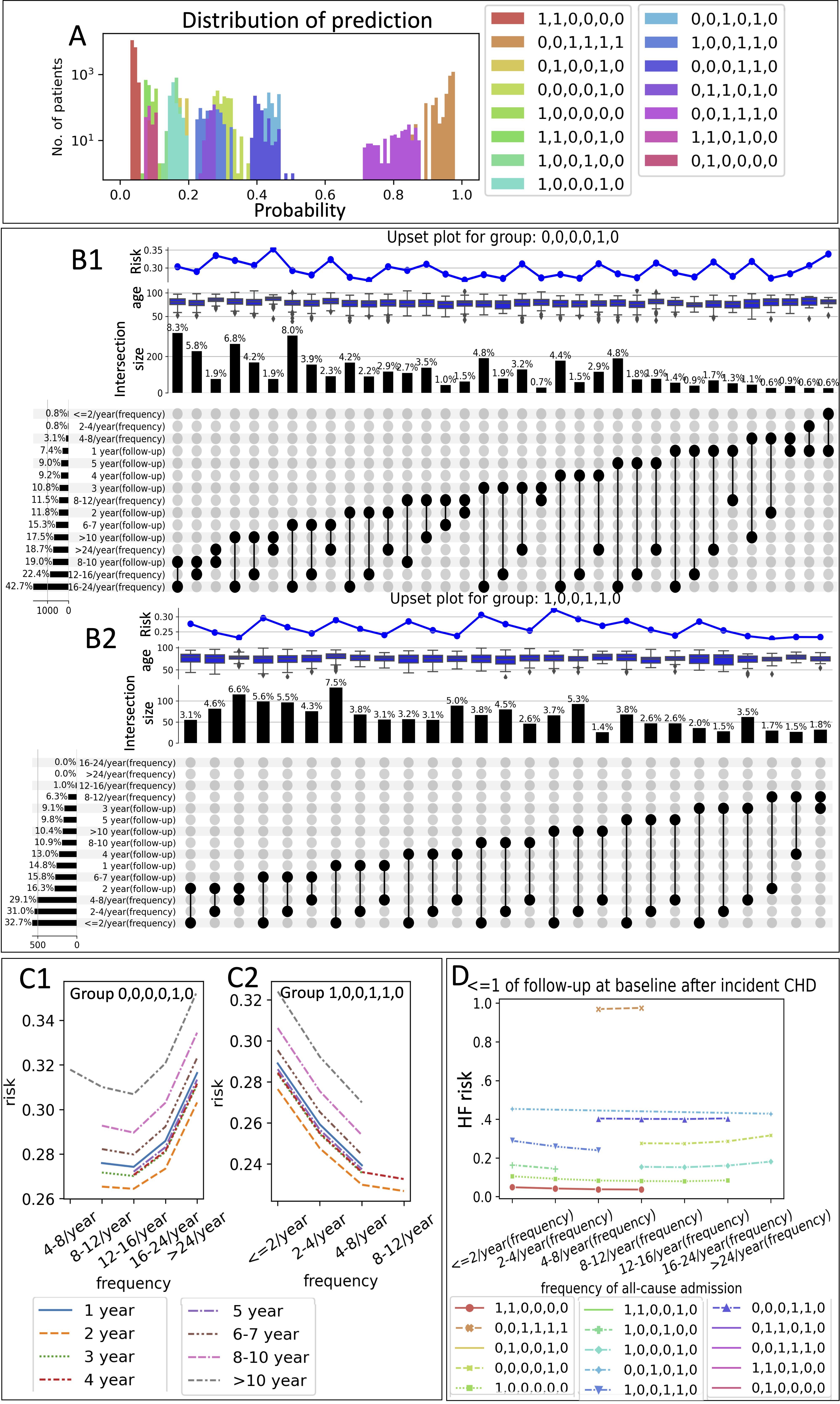}}
\vskip -0.1in
\caption{Descriptive analysis and counterfactual explanations for HF risk and frequency of all-cause visits at baseline after incident CHD stratified by discrete latent features. A: distribution of HF risk prediction on the validation set; the colours represent the patient clusters. B1 and B2 are UpSet plots of two example clusters for illustration. C1 and C2 summaries B1 and B2, showing the relationship among HF risk, frequency of all-cause admission, and follow-up in year at baseline for plausible counterfactuals in group “0,0,0,0,1,0” and “1,0,0,1,1,0”, respectively. The colours represent the follow-up in year after incidence CHD at baseline. D: relation between frequency of admission and HF risk for patients at the onset ($\leq$1 year) of CHD.}
\label{fig:f-hf}
\end{center}
\vskip -0.3in
\end{figure}

In this experiment, we investigated how having a different frequency of all-cause hospital and GP visits after incident CHD can influence the risk of HF. As shown in Figure~\ref{fig:f-hf}A, PCB maps CHD patients into 15 clusters, and different clusters can have different plausible counterfactuals. Two examples (Figure~\ref{fig:f-hf}B and C) were used to show that the frequency of all-cause admission can relate to HF risk differently for patients within different clusters. For example, Figure~\ref{fig:f-hf} C1 and C2 show HF risk increases and decreases with the increases of frequency of admission, in clusters “0,0,0,0,1,0” and “1,0,0,1,1,0”, respectively. However, both C1 and C2 show a consistent trend that a patient at the onset of CHD (i.e., less than a year) or diagnosed with CHD for more than five years can have a higher risk of HF than the other scenarios. 

Additionally, we investigated various plausible counterfactuals across different clusters to provide a population-wise overview using patients at the onset of CHD. Figure~\ref{fig:f-hf}D shows the frequency of all-cause admission relates to the HF risk differently across different clusters but in general, the HF risk decreases as the frequency increases from “$\leq$2/year” to “4-8/year”, and the HF risk increases as the frequency increases from “8-12/year” to “$>$24/year”. The UpSet plots for more clusters can be found in the Supplementary. 

\section{Discussion}
In this research, we proposed a PCB framework, which uses a combination of discrete latent representations (as auxiliary predictive and phenotyping features), and high-level human-understandable concepts (as intervention candidates). It achieved a similar prediction accuracy to its black-box counterpart, while providing counterfactual explanations. Furthermore, it avoided intensive expert-driven feature engineering, by focusing on untangling the relations between the outcome and the concepts of interest. By automatically phenotyping patients into different clusters using the latent representations, practitioners can investigate various counterfactuals for patients within a specific group to assist decision-making (e.g., provide potentially actionable suggestions), or get a better understanding of the model’s inner workings. 

In this study, we showcased two PCBs designs for HF risk prediction with an emphasis on different medical explanations: AF-HF and F-AF. Our analysis on the relation between HF and AF shows that AF can substantially increase the risk of HF in patients with diabetes and/or hypertension, but the RR of HF decreases as the HF risk of a cluster increases. This result indicates the importance of preventive efforts for AF across the trajectory of HF development, especially for patients at low or middle risk of HF. Additionally, our analysis shows that the HF risk is higher for patients with both diabetes and hypertension than those with only diabetes when exposed to AF. However, AF prevention in patients with only diabetes can be more beneficial than those with both diabetes and hypertension for HF prevention. This suggests that monitoring the prognosis of AF using ECG not only in patients with both diabetes and hypertension but also on patients with only diabetes can be a practical patient management strategy for HF prevention. Furthermore, the PCB model phenotypes patients who have only diabetes or hypertension as a comorbidity of AF into completely different clusters without intersection. This suggests that these two groups of patients can have different trajectories or underlying mechanisms that cause HF, hence, potentially requiring different prevention strategies. 
% although the absolute HF risk is slightly higher for patients with both diabetes and hypertension than those with only diabetes when exposed to AF, AF can considerably increase the risk of HF in patients with diabetes more than those with both diabetes and hypertension. 

Additionally, our analysis on the relation between HF and frequency of all-cause hospital and GP admission after incident CHD shows that the PCB model can phenotype heterogeneous population into different clusters and how the frequency of admission relates to the HF risk can differ across different clusters. Overall, our results suggest that the HF risk decreases as the frequency increases from “$\leq$2/year” to “4-8/year” and increases as the frequency increases from “8-12/year” to “$>$24/year”. Unlike AF, which consistently increases the HF risk across clusters, the relation between frequency of all-cause admission and HF risk is more diverse. It requires more careful analysis and better understanding about which cluster a patient belongs to assist decision-making or patient management.

A risk prediction model that experts can interact with to investigate what-if scenarios can be a natural fit and might be the most practical tool to assist decision-making. However, most of the previous studies usually consider risk prediction and model interpretability as two individual components. They focused on delivering accurate risk prediction first, then conducting post-hot interpretation using, for example, perturbation-based methods~\cite{guan2019towards}, saliency maps~\cite{smilkov2017smoothgrad}, and local or global surrogate methods~\cite{ribeiro2016should,vstrumbelj2014explaining,elshawi2019interpretability}. Without expert knowledge as guidance during model training, it is difficult to drive the post-hoc interpretation towards a specific question. The best interpretation these methods can provide is the association between a feature and the outcome based on the observations. Despite their usefulness in generating causal hypotheses to guide further clinical investigations, the interpretations fail to assist clinical decision-making or provide actionable suggestions. These can be more important and desired properties in risk prediction. 

On the contrary, epidemiologists have a particular interest in understanding the what-if question for a well-specified causal claim: How an exposure A causes the outcome B to change. This is achieved by conducting a randomized clinical trial (RCT) or emulating RCTs using observational data. Although common causal frameworks secure the interpretation of the results, they are not suitable for delivering a precise risk estimation on the general population. One of the major contributions of our work is that we embedded the causal framework within the risk prediction model and proposed a potential solution to fill the gap between prediction and causality. The proposed model can provide personalized risk prediction and counterfactual explanations (based on the cluster a patient belongs to). Additionally, instead of considering the concept of causality as emulating the RCTs, we proposed to think of causation in a pluralistic perspective~\cite{vandenbroucke2016causality} and used the empirical data to inform the possibility and plausibility of counterfactuals (i.e., the law of large numbers). This framework can not only answer our question regarding one exposure, but also can be as flexible as necessary to solve more complex and multi-dimensional questions of interest. 

Nevertheless, this study also has several limitations. More hyperparameter searching for the PCB models could have been considered, even though the reported models have achieved reasonably high performance and can lead to reasonable counterfactual analysis. Furthermore, the concepts of interest and latent representation are not independent. Thus, the descriptive analysis (i.e., UpSet plot) on the plausibility of counterfactuals is mandatory, and not all counterfactuals are plausible to deliver explanations. Another limitation is that cluster phenotyping is achieved by latent representation learning, which makes the mechanisms for phenotyping relatively difficult to explain. To alleviate this, for instance, a descriptive analysis of patients’ baseline characteristics within each cluster can be conducted to gain more insights and profile the patient groups.   

A model that can achieve accurate HF risk prediction and with the ability to provide what-if explanations is highly desired for disease prevention and patient management. By deploying such a model, physicians can assess a patient’s current HF risk and hypothetically investigate how different actions can influence the risk of HF and take the action that can benefit the outcome the most. With the growing availability of EHRs, we believe that the proposed model has a great potential to be integrated into routine care to provide guidance to assist decision-making. Although this study concerns HF, it can be applied to other questions related to HF as well as other outcomes.

\nocite{langley00}

\bibliography{example_paper}

\begin{thebibliography}{29}
\providecommand{\natexlab}[1]{#1}
\providecommand{\url}[1]{\texttt{#1}}
\expandafter\ifx\csname urlstyle\endcsname\relax
  \providecommand{\doi}[1]{doi: #1}\else
  \providecommand{\doi}{doi: \begingroup \urlstyle{rm}\Url}\fi

\bibitem[Baevski et~al.(2020)Baevski, Schneider, and
  Auli]{Baevski2020vq-wav2vec:}
Baevski, A., Schneider, S., and Auli, M.
\newblock vq-wav2vec: Self-supervised learning of discrete speech
  representations.
\newblock In \emph{International Conference on Learning Representations}, 2020.

\bibitem[Chisholm(1990)]{chisholm1990read}
Chisholm, J.
\newblock The read clinical classification.
\newblock \emph{BMJ: British Medical Journal}, 300\penalty0 (6732):\penalty0
  1092, 1990.

\bibitem[Committee et~al.(2014)]{jointbritish}
Committee, J.~F. et~al.
\newblock \emph{British National Formulary}.
\newblock BMJ Group and Pharmaceutical Press, 2014.

\bibitem[Conrad et~al.(2018)Conrad, Judge, Tran, Mohseni, Hedgecott, Crespillo,
  Allison, Hemingway, Cleland, McMurray, et~al.]{conrad2018temporal}
Conrad, N., Judge, A., Tran, J., Mohseni, H., Hedgecott, D., Crespillo, A.~P.,
  Allison, M., Hemingway, H., Cleland, J.~G., McMurray, J.~J., et~al.
\newblock Temporal trends and patterns in heart failure incidence: a
  population-based study of 4 million individuals.
\newblock \emph{The Lancet}, 391\penalty0 (10120):\penalty0 572--580, 2018.

\bibitem[Devlin et~al.(2019)Devlin, Chang, Lee, and Toutanova]{DBLP}
Devlin, J., Chang, M.-W., Lee, K., and Toutanova, K.
\newblock Bert: Pre-training of deep bidirectional transformers for language
  understanding.
\newblock In \emph{NAACL-HLT (1)}, pp.\  4171--4186, 2019.

\bibitem[Elshawi et~al.(2019)Elshawi, Al-Mallah, and
  Sakr]{elshawi2019interpretability}
Elshawi, R., Al-Mallah, M.~H., and Sakr, S.
\newblock On the interpretability of machine learning-based model for
  predicting hypertension.
\newblock \emph{BMC medical informatics and decision making}, 19\penalty0
  (1):\penalty0 1--32, 2019.

\bibitem[Goodfellow et~al.(2016)Goodfellow, Bengio, and
  Courville]{Goodfellow-et-al-2016}
Goodfellow, I., Bengio, Y., and Courville, A.
\newblock \emph{Deep Learning}.
\newblock MIT Press, 2016.

\bibitem[Goyal et~al.(2019)Goyal, Feder, Shalit, and Kim]{goyal2019explaining}
Goyal, Y., Feder, A., Shalit, U., and Kim, B.
\newblock Explaining classifiers with causal concept effect (cace).
\newblock \emph{arXiv preprint arXiv:1907.07165}, 2019.

\bibitem[Guan et~al.(2019)Guan, Wang, Zhang, Chen, He, and
  Xie]{guan2019towards}
Guan, C., Wang, X., Zhang, Q., Chen, R., He, D., and Xie, X.
\newblock Towards a deep and unified understanding of deep neural models in
  nlp.
\newblock In \emph{International Conference on Machine Learning}, pp.\
  2454--2463. PMLR, 2019.

\bibitem[Herrett et~al.(2015)Herrett, Gallagher, Bhaskaran, Forbes, Mathur,
  Van~Staa, and Smeeth]{herrett2015data}
Herrett, E., Gallagher, A.~M., Bhaskaran, K., Forbes, H., Mathur, R., Van~Staa,
  T., and Smeeth, L.
\newblock Data resource profile: clinical practice research datalink (cprd).
\newblock \emph{International Journal of Epidemiology}, 44\penalty0
  (3):\penalty0 827--836, 2015.

\bibitem[Koh et~al.(2020)Koh, Nguyen, Tang, Mussmann, Pierson, Kim, and
  Liang]{koh2020concept}
Koh, P.~W., Nguyen, T., Tang, Y.~S., Mussmann, S., Pierson, E., Kim, B., and
  Liang, P.
\newblock Concept bottleneck models.
\newblock In \emph{International Conference on Machine Learning}, pp.\
  5338--5348. PMLR, 2020.

\bibitem[Kuan et~al.(2019)Kuan, Denaxas, Gonzalez-Izquierdo, Direk, Bhatti,
  Husain, Sutaria, Hingorani, Nitsch, Parisinos, et~al.]{kuan2019chronological}
Kuan, V., Denaxas, S., Gonzalez-Izquierdo, A., Direk, K., Bhatti, O., Husain,
  S., Sutaria, S., Hingorani, M., Nitsch, D., Parisinos, C.~A., et~al.
\newblock A chronological map of 308 physical and mental health conditions from
  4 million individuals in the english national health service.
\newblock \emph{The Lancet Digital Health}, 1\penalty0 (2):\penalty0 e63--e77,
  2019.

\bibitem[Lewis(2013)]{lewis2013counterfactuals}
Lewis, D.
\newblock \emph{Counterfactuals}.
\newblock John Wiley \& Sons, 2013.

\bibitem[Lex et~al.(2014)Lex, Gehlenborg, Strobelt, Vuillemot, and
  Pfister]{lex2014upset}
Lex, A., Gehlenborg, N., Strobelt, H., Vuillemot, R., and Pfister, H.
\newblock Upset: visualization of intersecting sets.
\newblock \emph{IEEE Transactions on Visualization and Computer Graphics},
  20\penalty0 (12):\penalty0 1983--1992, 2014.

\bibitem[Li et~al.(2016)Li, Chen, Hovy, and Jurafsky]{li-etal-2016-visualizing}
Li, J., Chen, X., Hovy, E., and Jurafsky, D.
\newblock Visualizing and understanding neural models in {NLP}.
\newblock In \emph{Proceedings of the 2016 Conference of the North {A}merican
  Chapter of the Association for Computational Linguistics: Human Language
  Technologies}, pp.\  681--691, 2016.

\bibitem[Li et~al.(2021)Li, Mamouei, Salimi-Khorshidi, Rao, Hassaine, Canoy,
  Lukasiewicz, and Rahimi]{li2021hi}
Li, Y., Mamouei, M., Salimi-Khorshidi, G., Rao, S., Hassaine, A., Canoy, D.,
  Lukasiewicz, T., and Rahimi, K.
\newblock Hi-behrt: Hierarchical transformer-based model for accurate
  prediction of clinical events using multimodal longitudinal electronic health
  records.
\newblock \emph{arXiv preprint arXiv:2106.11360}, 2021.

\bibitem[Molnar(2019)]{molnar2019}
Molnar, C.
\newblock \emph{Interpretable Machine Learning}.
\newblock 2019.

\bibitem[Organization(2004)]{world2004international}
Organization, W.~H.
\newblock \emph{The International Statistical Classification of Diseases and
  Health Related Problems ICD-10: Tenth Revision. Volume 1: Tabular List},
  volume~1.
\newblock World Health Organization, 2004.

\bibitem[{O\textquotesingle Shaughnessy} et~al.(2020){O\textquotesingle
  Shaughnessy}, Canal, Connor, Rozell, and Davenport]{NEURIPS2020_3a93a609}
{O\textquotesingle Shaughnessy}, M., Canal, G., Connor, M., Rozell, C., and
  Davenport, M.
\newblock {Generative causal explanations of black-box classifiers}.
\newblock In Larochelle, H., Ranzato, M., Hadsell, R., Balcan, M.~F., and Lin,
  H. (eds.), \emph{Advances in Neural Information Processing Systems},
  volume~33, pp.\  5453--5467, 2020.

\bibitem[Pearl(2009)]{pearl2009causality}
Pearl, J.
\newblock \emph{Causality}.
\newblock Cambridge university press, 2009.

\bibitem[Pearl \& Mackenzie(2018)Pearl and Mackenzie]{pearl2018book}
Pearl, J. and Mackenzie, D.
\newblock \emph{The book of why: the new science of cause and effect}.
\newblock Basic books, 2018.

\bibitem[Ribeiro et~al.(2016)Ribeiro, Singh, and Guestrin]{ribeiro2016should}
Ribeiro, M.~T., Singh, S., and Guestrin, C.
\newblock " why should i trust you?" explaining the predictions of any
  classifier.
\newblock In \emph{Proceedings of the 22nd ACM SIGKDD International Conference
  on Knowledge Discovery and Data Mining}, pp.\  1135--1144, 2016.

\bibitem[Smilkov et~al.(2017)Smilkov, Thorat, Kim, Vi{\'e}gas, and
  Wattenberg]{smilkov2017smoothgrad}
Smilkov, D., Thorat, N., Kim, B., Vi{\'e}gas, F., and Wattenberg, M.
\newblock Smoothgrad: removing noise by adding noise.
\newblock \emph{arXiv preprint arXiv:1706.03825}, 2017.

\bibitem[{\v{S}}trumbelj \& Kononenko(2014){\v{S}}trumbelj and
  Kononenko]{vstrumbelj2014explaining}
{\v{S}}trumbelj, E. and Kononenko, I.
\newblock Explaining prediction models and individual predictions with feature
  contributions.
\newblock \emph{Knowledge and Information Systems}, 41\penalty0 (3):\penalty0
  647--665, 2014.

\bibitem[Szklo \& Nieto(2014)Szklo and Nieto]{szklo2014epidemiology}
Szklo, M. and Nieto, F.~J.
\newblock \emph{Epidemiology: beyond the basics}.
\newblock Jones \& Bartlett Publishers, 2014.

\bibitem[van~den Oord et~al.(2017)van~den Oord, Vinyals, and koray
  Kavukcuoglu]{NIPS2017_7a98af17}
van~den Oord, A., Vinyals, O., and koray Kavukcuoglu.
\newblock {Neural Discrete Representation Learning}.
\newblock In Guyon, I., Luxburg, U.~V., Bengio, S., Wallach, H., Fergus, R.,
  Vishwanathan, S., and Garnett, R. (eds.), \emph{Advances in Neural
  Information Processing Systems}, volume~30, 2017.

\bibitem[Vandenbroucke et~al.(2016)Vandenbroucke, Broadbent, and
  Pearce]{vandenbroucke2016causality}
Vandenbroucke, J.~P., Broadbent, A., and Pearce, N.
\newblock Causality and causal inference in epidemiology: the need for a
  pluralistic approach.
\newblock \emph{International Journal of Epidemiology}, 45\penalty0
  (6):\penalty0 1776--1786, 2016.

\bibitem[Yang \& Kim(2019)Yang and Kim]{BAM2019}
Yang, M. and Kim, B.
\newblock {Benchmarking Attribution Methods with Relative Feature Importance}.
\newblock \emph{CoRR}, abs/1907.09701, 2019.

\bibitem[Yoon et~al.(2018)Yoon, Jordon, and van~der Schaar]{yoon2018invase}
Yoon, J., Jordon, J., and van~der Schaar, M.
\newblock Invase: Instance-wise variable selection using neural networks.
\newblock In \emph{International Conference on Learning Representations}, 2018.

\end{thebibliography}
\bibliographystyle{icml2021}

\newpage
\beginsupplement
\section{Supplementary Contents}
Supplementary Methods and Discussion\\
1.	Data processing for Hi-BEHRT\\
2.	Hi-BEHRT\\
3.	Experimental details\\
4.	Concept accuracy\\
5.  Sanity check for counterfactual analysis

Supplementary Tables\\
Table S1. ICD-10 codes used to identify patients with heart failure in hospital discharge records and general practice records\\
Table S2. Descriptive analysis of HF cohort

Supplementary Figures\\
Figure S1. Hi-BEHRT model architecture\\
Figure S2. Histograms of F1 score of individual concepts averaged over multiple random seeds\\
Figure S3. Observed risk (prevalence) ratio of HF for AF-HF task\\
Figure S4. UpSet plots for patients with only diabetes or hypertension\\
Figure S5. UpSet plots for F-HF\\

\section{Supplementary Methods and Discussion}
\subsection{Data processing for Hi-BEHRT}
We included records from diagnoses using International classification of diseases, tenth revision (ICD-10)~\cite{world2004international} codes in level four; medications using British National Formulary coding scheme~\cite{jointbritish} in the section level, GP tests in Read code~\cite{chisholm1990read}, and hospital procedures using The Office of Population Censuses and Surveys (OPCS) Classification of Interventions and Procedures codes, blood pressure (BP) measurement (both systolic and diastolic pressure) in mmHg, drinking status, smoking status and body mass index (BMI) in $kg/m^2$ for modelling. Both drinking status and smoking status were recorded as categorical value, including current drinker/smoker, ex, and non. For continuous values, more specifically, BP and BMI, we categorised their value by 5 $mmHg$ and 1 $kg/m^2$, respectively, and excluded abnormal values: higher than 200 and lower than 80 $mmHg$ for systolic pressure, higher than 140 and lower than 50 $mmHg$ for diastolic pressure, and high than 50 and lower than 16 $kg/m^2$ for BMI. Additionally, we also calculated the corresponding age for each record using event date and date of birth for the convenience of modelling.

The incidence of HF was defined as the first record of HF recorded in the primary care or hospital admission records. We considered HF as a composite condition of rheumatic heart failure, hypertensive heart and disease with (congestive) heart failure and renal failure, ischemic cardiomyopathy, chronic cor pulmonale, congestive heart failure, cardiomyopathy, left ventricular failure, and cardiac, heart, or myocardial failure~\cite{conrad2018temporal}. 

The HF risk prediction task in this work was defined as using EHRs before the baseline to predict the risk of HF within 5 years after the baseline. To address the concerns on the inaccuracy of time between the date of event recording and the date of event occurrence, we explicitly ignored records at least one year before the incidence of HF. Therefore, the baseline was selected as a random date between one year and five years before the incidence of HF for HF (+) patients. For HF (-) patients, the baseline is selected as a random date within a patient’s medical history if there were five-year complete records after the selected baseline to ensure a patient is free of HF. Afterwards, we only included patients who are registered with GPs for more than three years before the baseline.

\subsection{Hi-BEHRT}
Hi-BEHRT~\cite{li2021hi} is a Transformer-based hierarchical risk prediction model. It predicts outcome by incorporating a patient’s complete medical history. Similar to Transformer-based model, Hi-BEHRT takes medical records (diagnoses, medications, and records from other sources) as input, with corresponding age, segmentation, and position code as auxiliary features to indicate the sequential order of medical records. Age represents the age of a patient when an event is recorded, segmentation is a symbol to separate records between visits, and position code is a popular technique used for Transformer to indicate sequence order~\cite{DBLP}. Afterwards, a hierarchical Transformer, which contains a feature extractor and a feature aggregator, is used for risk prediction.  The feature extractor uses Transformer as a sliding window to extract local representation for only a fraction of records (segment) a time, and the representation of the first timestep in the last layer is extracted as the representation for this segment. The feature aggregator then uses a Transformer to aggregate segment representations to carry out the final prediction. This hierarchical architecture intends to address a limitation of Transformer on dealing with long sequence. The model architecture can be found in Figure S1.

\subsection{Experimental details}
\subsubsection{Standard black-box model (Hi-BEHRT)}
We adopted Hi-BEHRT and its hyper-parameter setup for CVD-HF, F-HF, and AF-HF. More specifically, it has 4 layers of feature extractor and 4 layers of feature aggregator, hidden dimension 150, number of attention heads 6, intermediate size 108, dropout rate 0.2, and attention dropout rate 0.3, maximum sequence length 1220, and 50 and 30 as window size and stride size, respectively, for the sliding window mechanism. We trained on 2 GPUs with a batch size of 128 per GPU giving a total batch size of 256. The model is trained for 100 epochs with 10\%, 40\%, and 50\% for warm-up, hold, and cosine decay, respectively, and early stop is applied when loss does not decrease for 10 epochs.  The learning rate for hold stage is $1e^{-4}$.

\subsubsection{AF-HF}
We used the identical Hi-BEHRT as mentioned above for latent representation learning. However, instead of pooling the first-time step for classification, we used a two-layer multi-layer perceptron to map the representation to the high-level concepts with 64 units for the first layer. The second layer has the number of units that is equivalent to the number of concepts. For vector quantization, we used binary variable to represent each dimension of the discrete latent representation. We searched on the number of dimensions for the latent representation over [4, 6]. Because each dimension is a binary variable, we used codebook to map each variable to the embedding space with just one dimension. A three-layer multi-layer perceptron with units 16, 8, 1 was followed for HF risk prediction.

\subsubsection{F-HF}
F-HF mapped the frequency of hospital and GP admission into 7 categories, $<=$2 (times)/year, 2-4/year, 4-8/year, 8-12/year, 12-16/year, 16-24/year, $>$24/year, and mapped the follow-up in year at baseline after incident CHD into 8 categories: 0-1 (year), 1-2, 2-3, 3-4, 4-5, 5-7, 7-10, >10. Therefore, these two concepts are categorical variables and are mapped into 2-dimensional embedding space with two different embedding metrices.  The rest of the setups are the same as described in CVD-HF and AF-HF.

For partial concept bottleneck models, we set up the temperature scale of Gumbel-Softmax quantization with initial value 2 and minimum value 0.5 with decay rate 0.999. We trained on 2 GPUs with a batch size of 128 per GPU giving a total batch size of 256. The model is trained for 100 epochs with 10\%, 40\%, and 50\% for warm-up, hold, and cosine decay, respectively, and early stop is applied when loss does not decrease for 10 epochs.  The learning rate for hold stage is $5e^{-5}$.

\subsection{Concept accuracy}
As suggested by Koh et al.~\cite{koh2020concept}, Figure S2 further shows that the PCBs in general can accurately predict each concept of interest, and the binary concepts (AF-HF) have better accuracy than the categorical concepts (F-HF). It means the predicted concepts are aligned with the true concepts and suggests we might intervene on the concepts effectively to investigate the counterfactual explanations afterwards.

\subsection{Sanity check for counterfactual analysis}
In this work, we used the AF-HF as an example to conduct a sanity check for the counterfactual analysis. To this purpose, we compared the consistency between the estimated risk ratio and the observed risk ratio within each group. As mentioned in the paper, the partial concept bottleneck model maps patients into latent groups and patients within the same latent group share the same latent covariates. Thus, we would expect that for patients who are within the same latent group and are well-stratified by the latent covariates, the estimated risk ratio of HF should in general align with the observed risk ratio. The estimated risk ratio is calculated by the HF risk estimated by the risk model, and the observed risk ratio is calculated by the prevalence of the HF risk within the exposure and non-exposure group. As shown in Figure S3, the observed risk ratio shows consistent pattern as the estimated risk ratio, therefore, reassure the validity of the counterfactual analysis.

\onecolumn
\section{Supplementary Tables}
\begin{table}[h!]
\caption{ICD-10 codes used to identify patients with heart failure in hospital discharge records and general practice records}
% \label{tab:accuracy}
\vskip -0.15in
\begin{center}
\begin{small}
\begin{sc}
\begin{tabular}{|c|c|}
\toprule
 ICD Code&Description \\
\midrule
I09.9&	Rheumatic heart failure\\
\midrule
I11.0&	Hypertensive heart disease with (congestive) heart failure\\
\midrule
I13.0&	Hypertensive heart and renal disease with (congestive) heart failure\\
\midrule
I13.2&	Hypertensive heart and renal disease with both (congestive) heart failure and renal failure\\
\midrule
I25.5&	Ischemic cardiomyopathy\\
\midrule
I27.9&	Chronic cor pulmonale\\
\midrule
I38&	Congestive heart failure due to valvular disease \\
\midrule
I42.0&	Congestive cardiomyopathy\\
\midrule
I42.1&	Obstructive hypertrophic cardiomyopathy\\
\midrule
I42.2&	Nonobstructive hypertrophic cardiomyopathy\\
\midrule
I42.6&	Alcoholic cardiomyopathy\\
\midrule
I42.8&	Other cardiomyopathies\\
\midrule
I42.9&	Cardiomyopathy NOS\\
\midrule
I50.0&	Congestive heart failure\\
\midrule
I50.1&	Left ventricular failure\\
\midrule
I50.2&	Systolic (congestive) heart failure\\
\midrule
I50.3&	Diastolic (congestive) heart failure\\
\midrule
I50.8&	Other heart failure\\
\midrule
I50.9&	Cardiac, heart or myocardial failure NOS\\

\bottomrule
\end{tabular}
\end{sc}
\end{small}
\end{center}
\vskip -0.1in
\end{table}

\begin{table}[h!]
\caption{Descriptive analysis of HF cohort}
% \label{tab:accuracy}
\vskip 0.in
\begin{center}
\begin{small}
\begin{sc}
\begin{tabular}{|l|l|}
\toprule
\multicolumn{2}{|l|}{General characteristics}\\
\midrule
No. of patients	& 1,975,630\\
\midrule
No. (\%) of HF patients	& 94, 495 (4.7)\\
\midrule
Male (\%)	& 810,864 (41.0)\\
\midrule
\multicolumn{2}{|l|}{Baseline characteristics}\\
\midrule
Median (IQR) No. of visits per patient	& 38 (64)\\
\midrule
Median (IQR) No. of codes per visit	&3.92 (1.86)\\
\midrule
Mean (SD) learning period before baseline (year)&	8.4 (4.0)\\
\midrule
Median (IQR) baseline age (year)&	52 (29)\\
\midrule
Hypertension (\%)&	433,319 (21.9)\\
\midrule
Diabetes (\%)&	130,928 (6.6)\\
\midrule
Atrial fibrillation (\%)&	63,899 (3.2)\\
\midrule
CHD (\%)	&109,123 (5.5)\\
\midrule
\multicolumn{2}{|l|}{Additional characteristics for CHD patients}\\
\midrule
Median (IQR) follow-up (year) at baseline after incident CHD	&5 (6)\\
\midrule
Median (IQR) frequency of all-cause hospital and GP admission per year after incident CHD&	12.0 (12.7)\\
\bottomrule
\multicolumn{2}{c}{SD: standard deviation, IQR: interquartile range}\\
\multicolumn{2}{c}{learning period is the time period before the baseline that is used for learning}\\

\end{tabular}
\end{sc}
\end{small}
\end{center}
\vskip -0.1in
\end{table}

\newpage
\section{Supplementary Figures}
\begin{figure}[h]
% \vskip -0.2in
% \begin{center}
\centerline{\includegraphics[width=0.7\columnwidth]{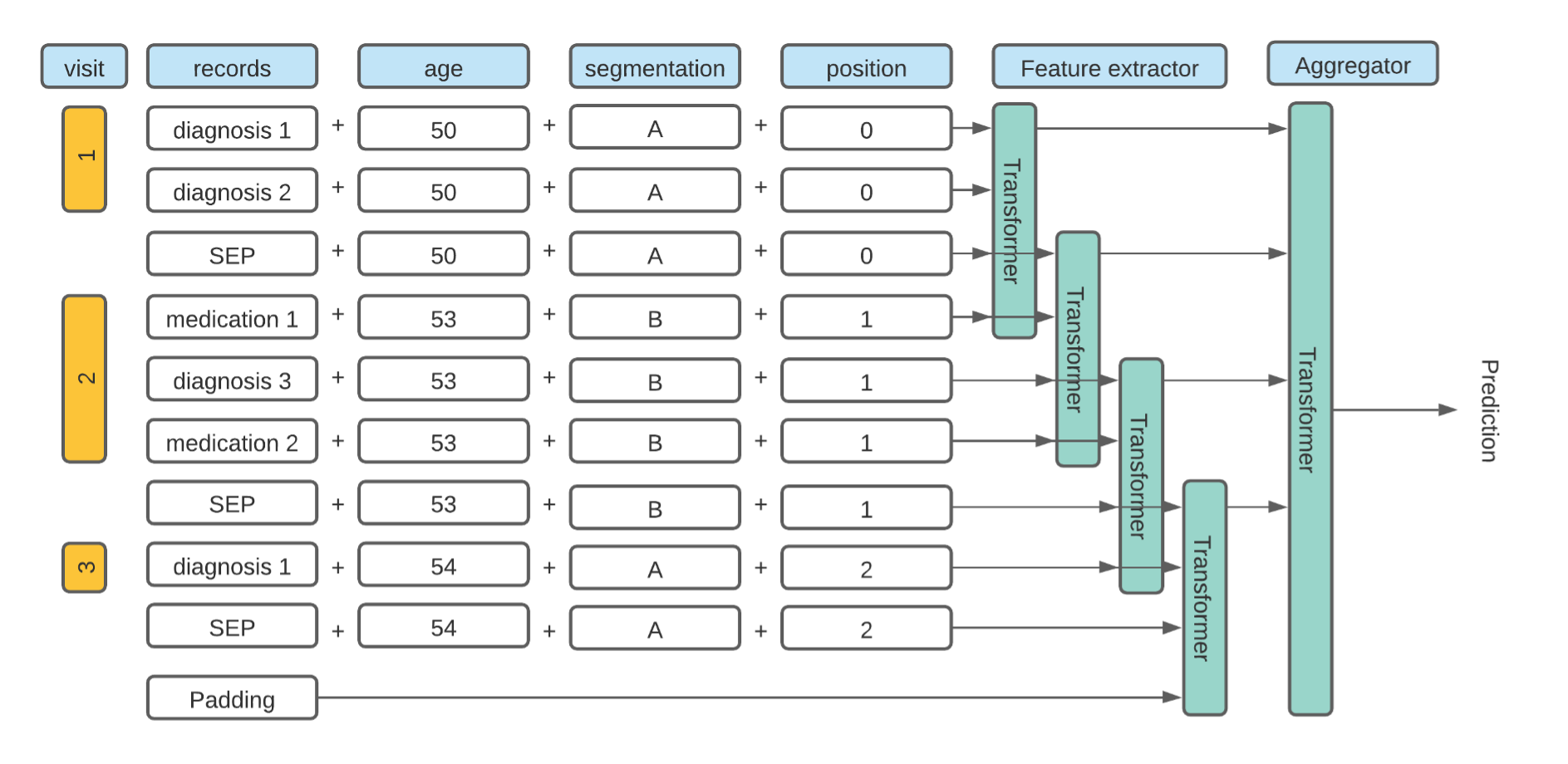}}
\caption{Hi-BEHRT model architecture. The representation of each encounter is the summation of presentations of records, age, segmentation, and position. }
\label{fig:hi-behrt}
% \end{center}
\vskip -0.2in
\end{figure}

\begin{figure}[h!]
% \vskip -0.2in
\begin{center}
\centerline{\includegraphics[width=0.5\columnwidth]{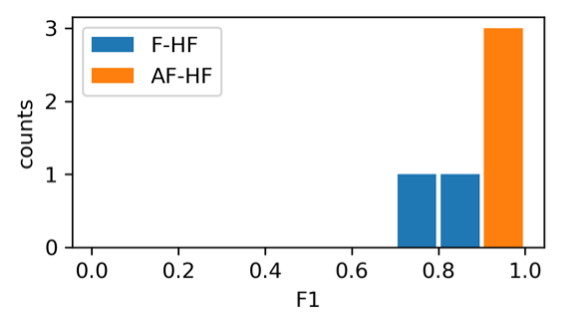}}
\caption{Histograms of F1 score of individual concepts averaged over multiple random seeds. The concepts of interest in different setups can be accurately predicted by PCBs.}
\label{fig:acc}
\end{center}
\vskip -0.2in
\end{figure}

\begin{figure}[h!]
% \vskip -0.2in
\begin{center}
\centerline{\includegraphics[width=0.5\columnwidth]{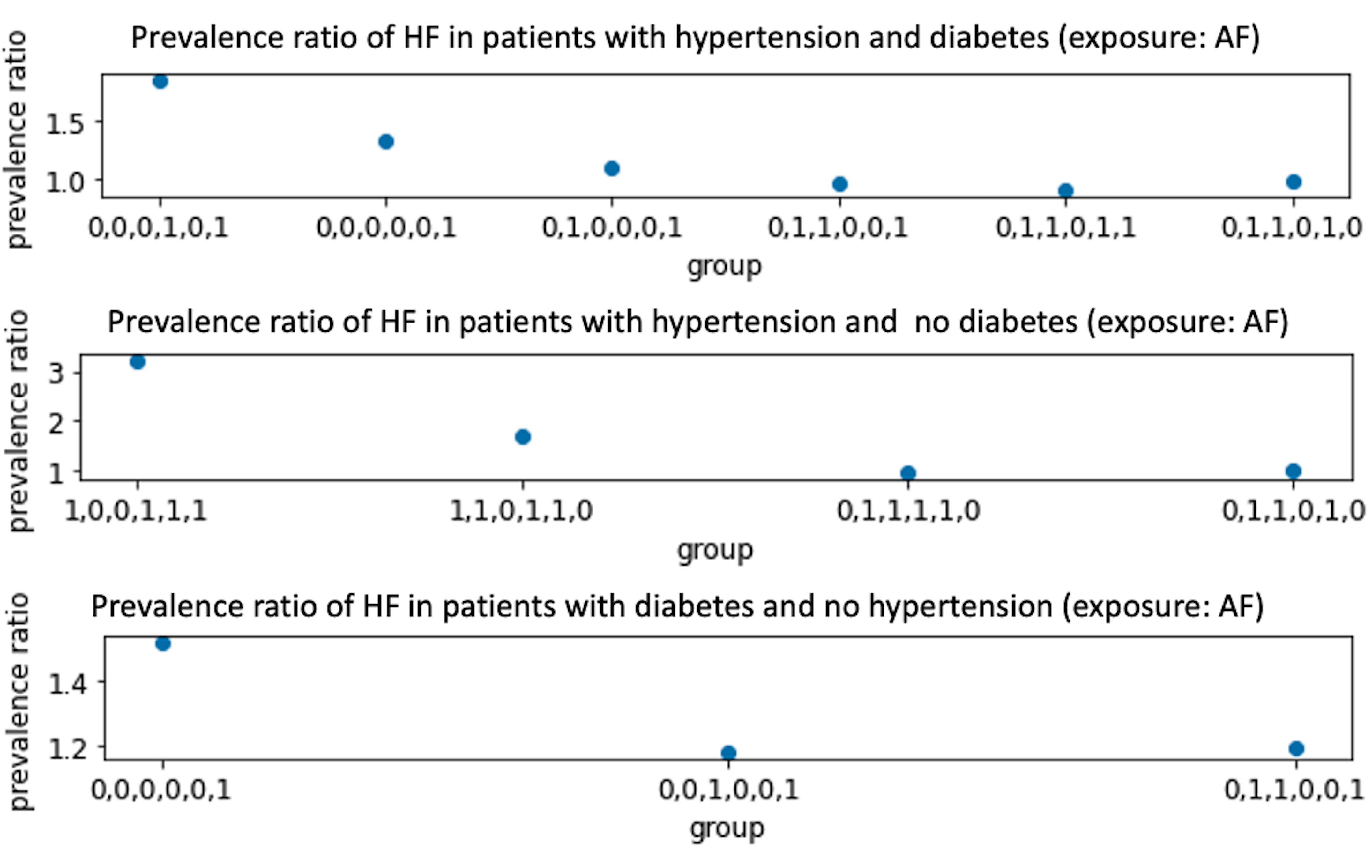}}
\caption{Observed risk (prevalence) ratio of HF for AF-HF task}
\label{fig:sanity}
\end{center}
\vskip -0.2in
\end{figure}

\begin{figure}[h!]
% \vskip -0.2in
\begin{center}
\centerline{\includegraphics[width=\columnwidth]{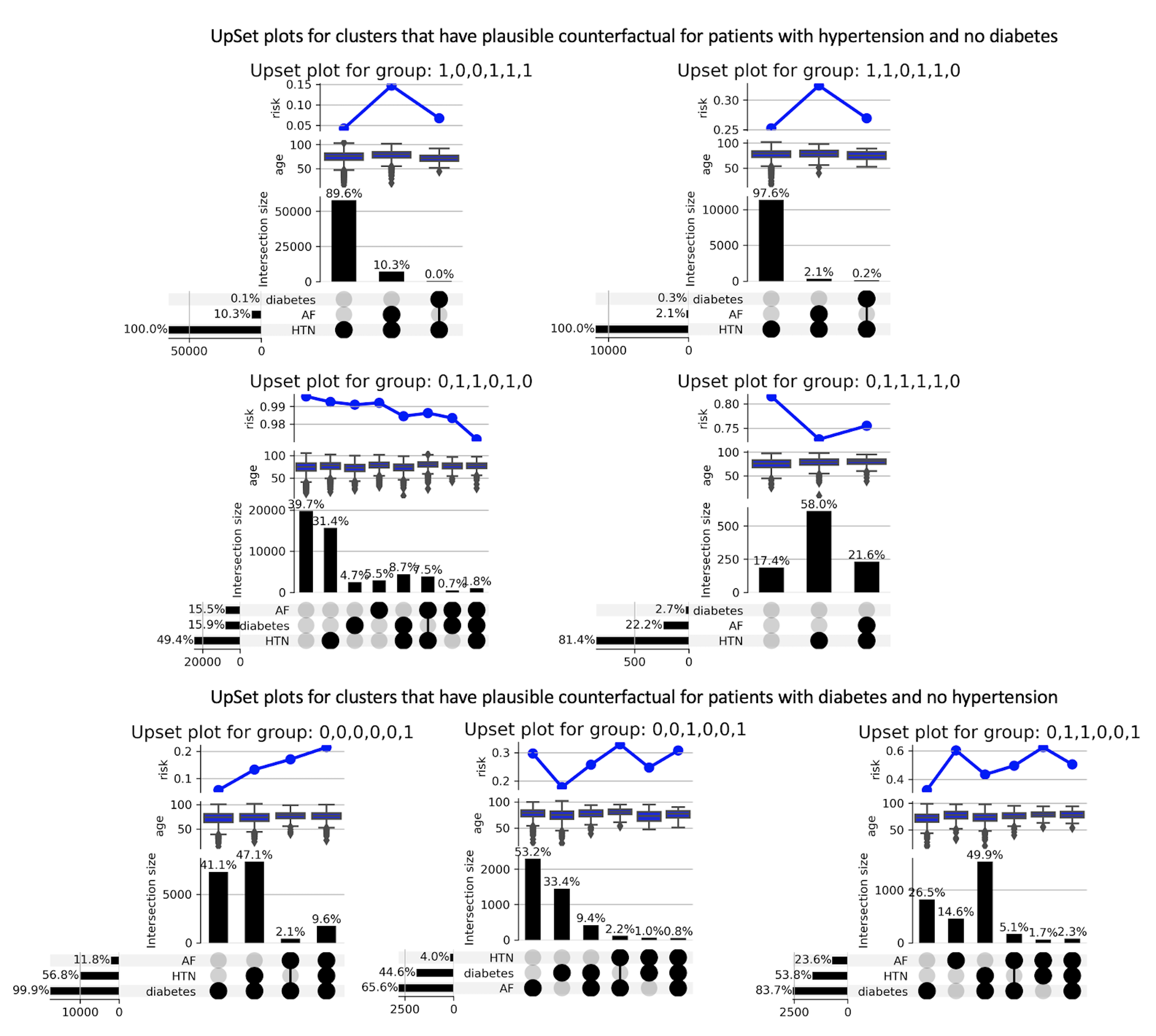}}
\caption{UpSet plots for patients with only diabetes or hypertension}
% \label{fig:acc}
\end{center}
\vskip -0.2in
\end{figure}

\begin{figure}[h!]
     \centering
         \includegraphics[width=\textwidth]{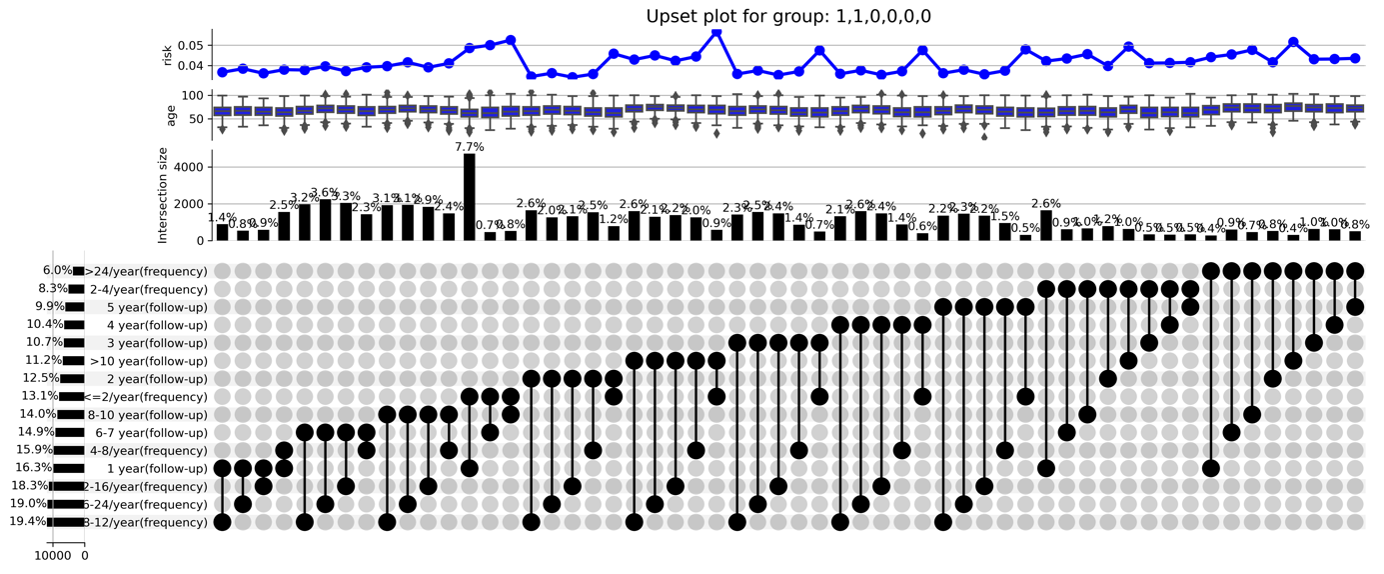}
         \centering
         \includegraphics[width=\textwidth]{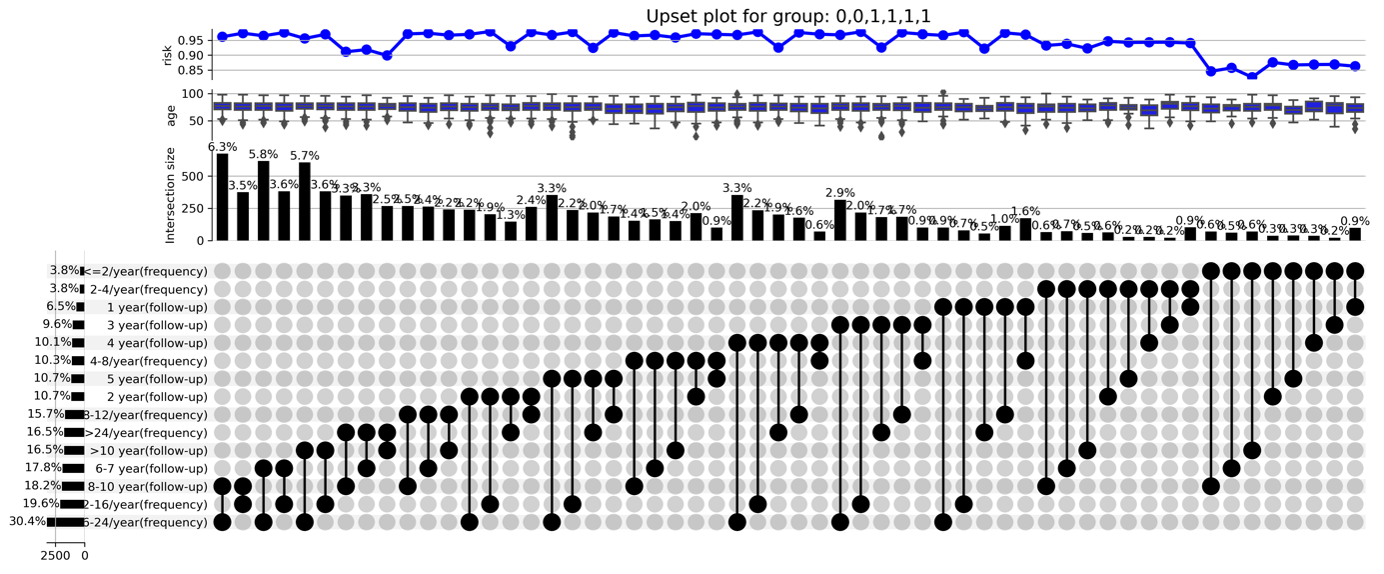}
 \centering
         \includegraphics[width=\textwidth]{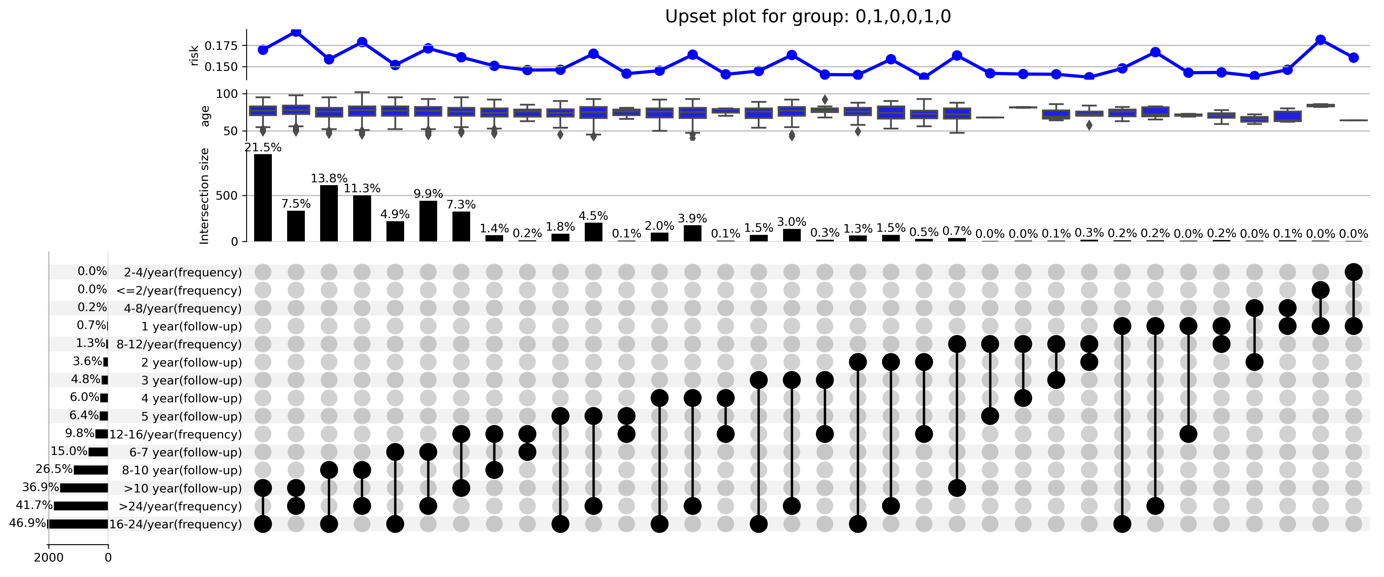}
\end{figure}
\begin{figure}[h!]
     \centering
         \includegraphics[width=\textwidth]{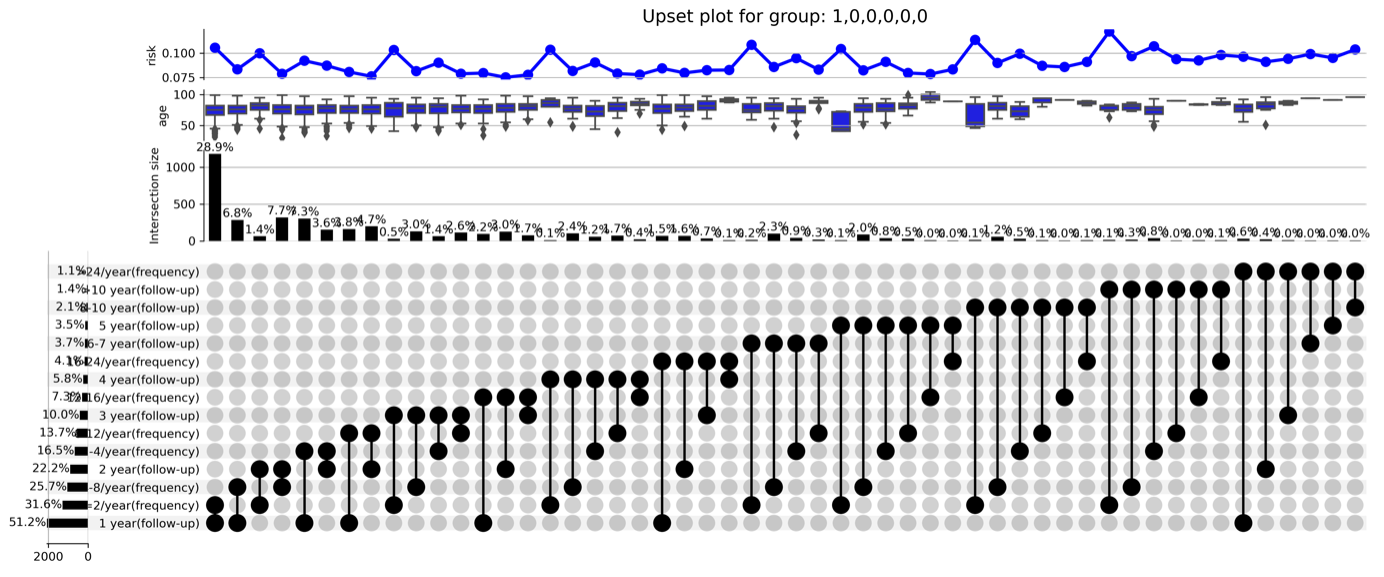}
         \centering
         \includegraphics[width=\textwidth]{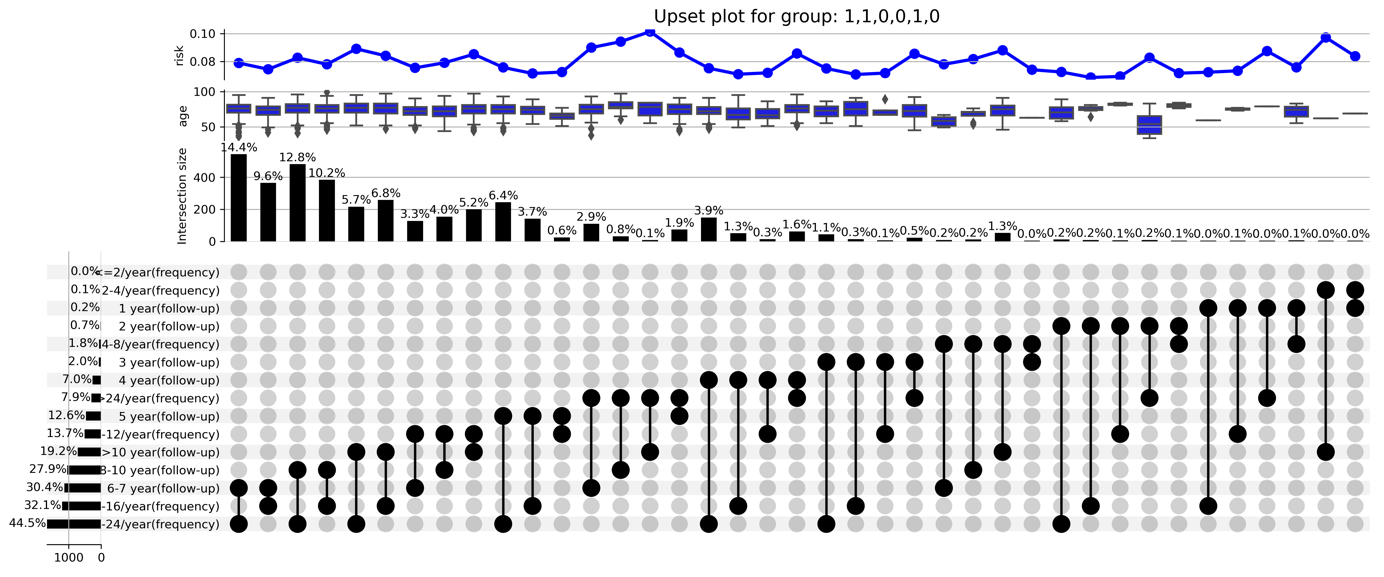}
 \centering
         \includegraphics[width=\textwidth]{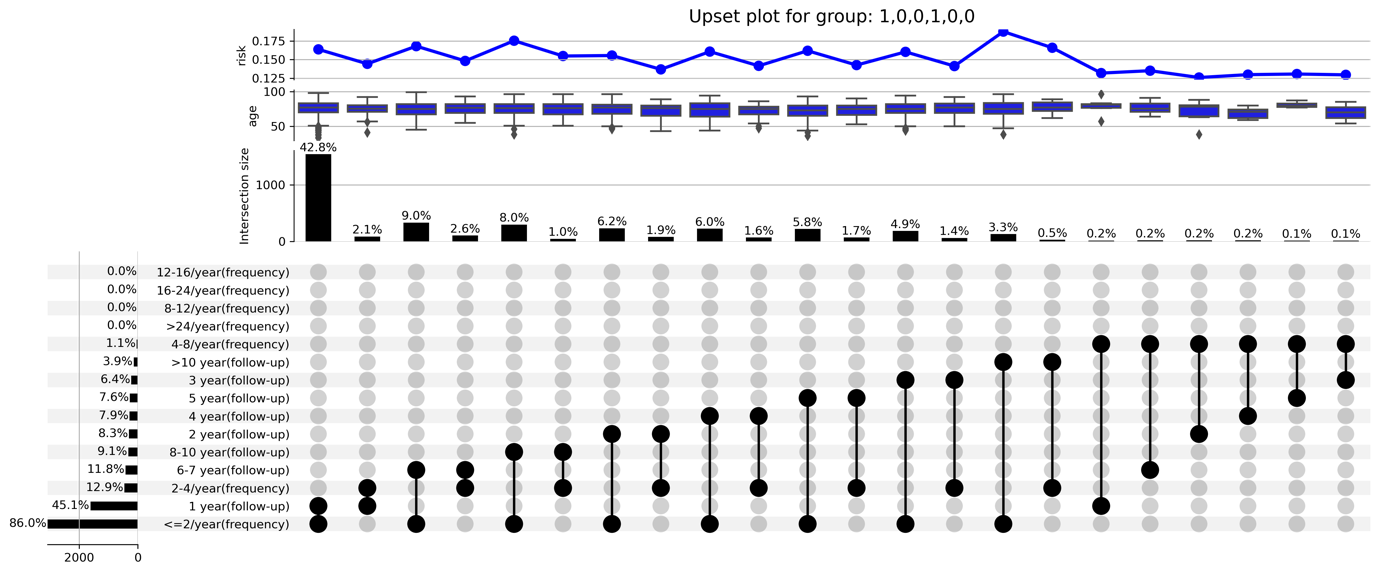}
\end{figure}
\begin{figure}[h!]
     \centering
         \includegraphics[width=\textwidth]{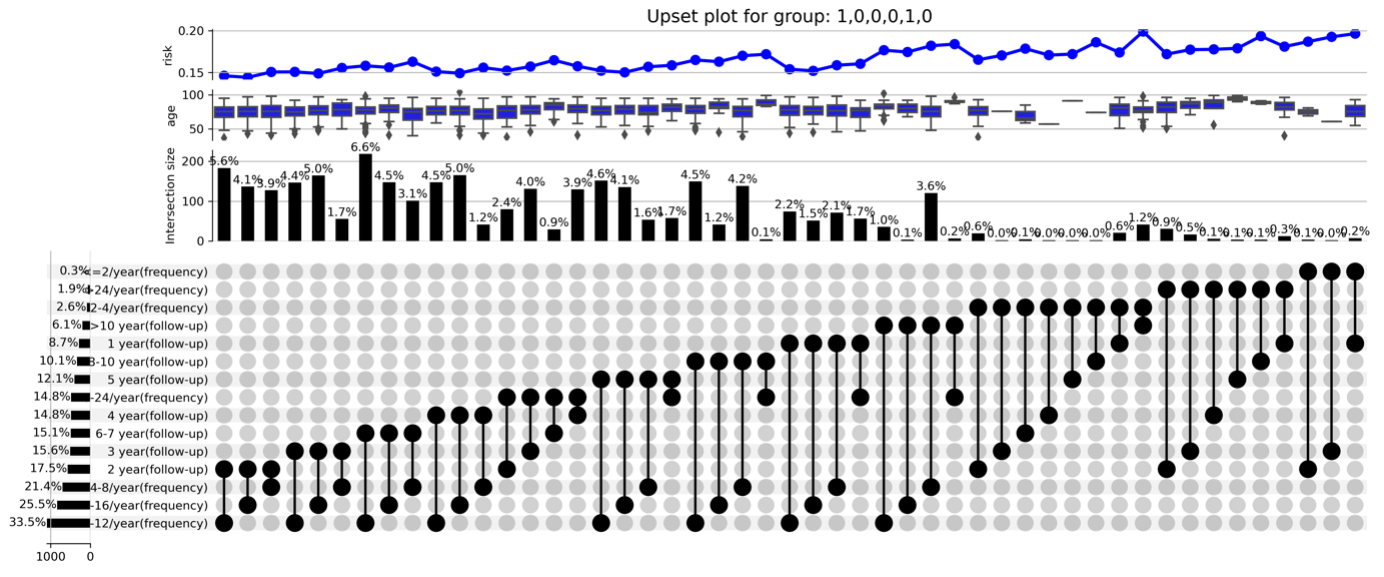}
         \centering
         \includegraphics[width=\textwidth]{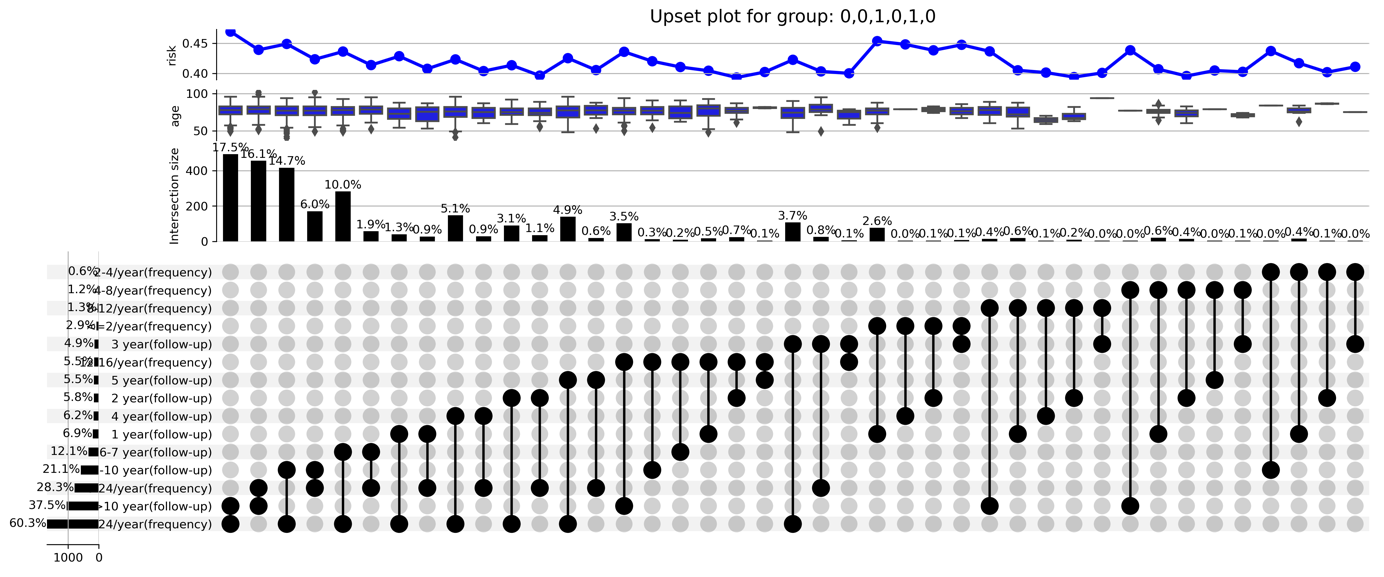}
 \centering
         \includegraphics[width=\textwidth]{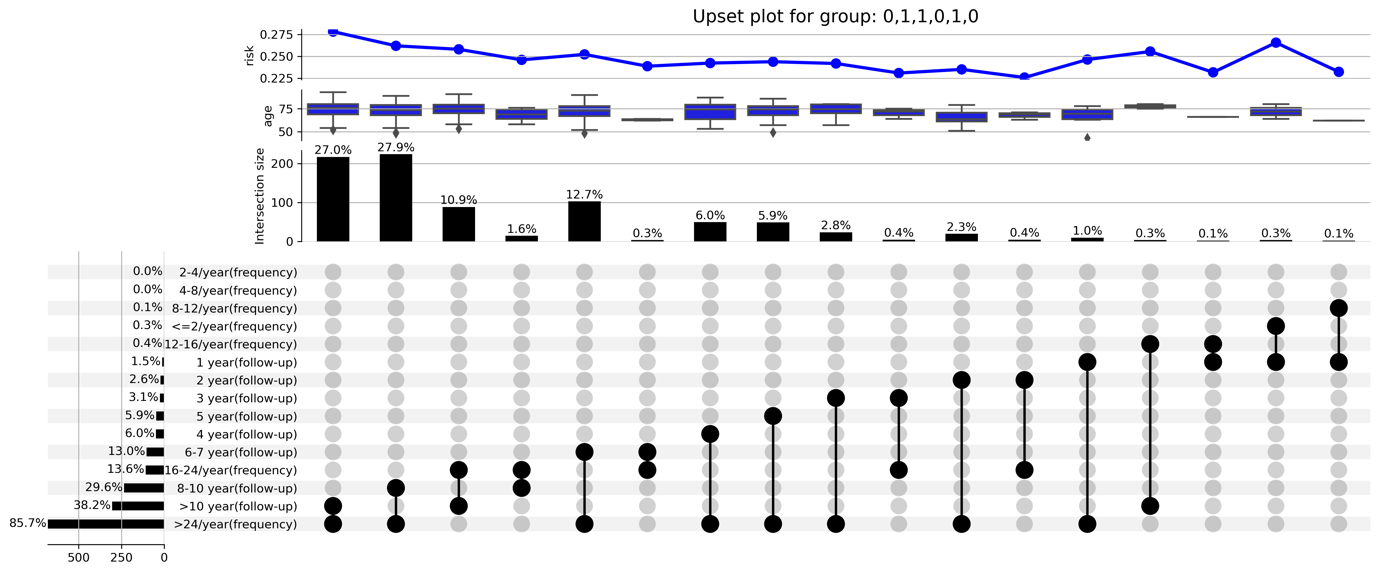}
\end{figure}
\begin{figure}[h!]
     \centering
         \includegraphics[width=\textwidth]{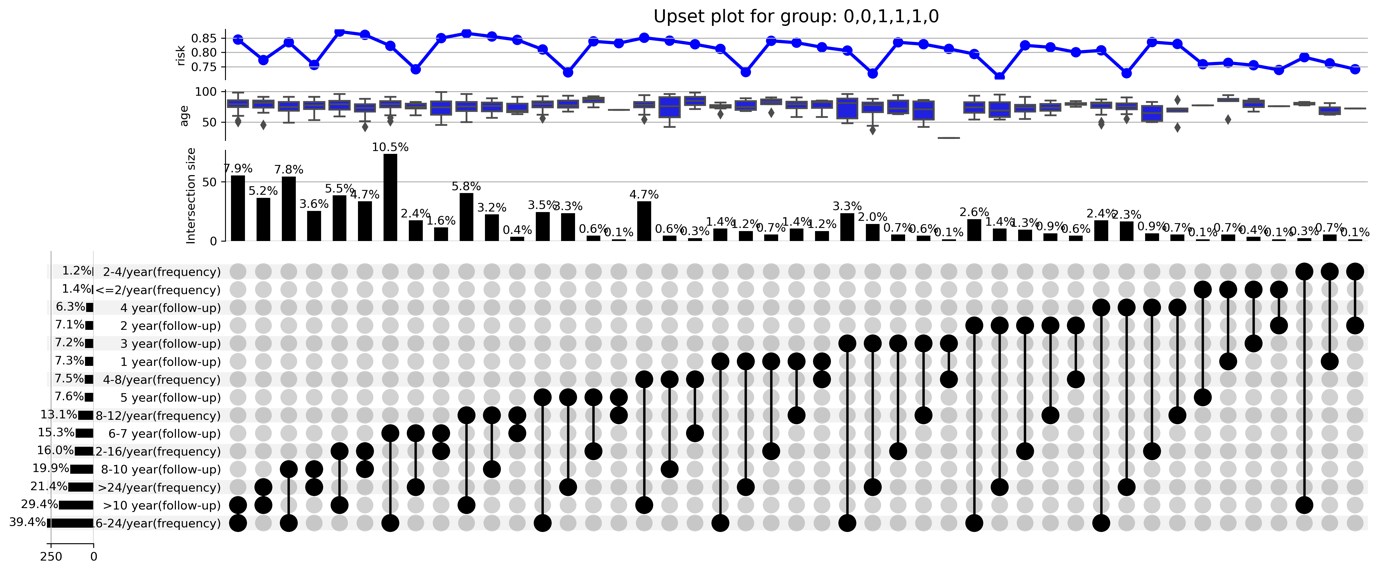}
         \centering
         \includegraphics[width=\textwidth]{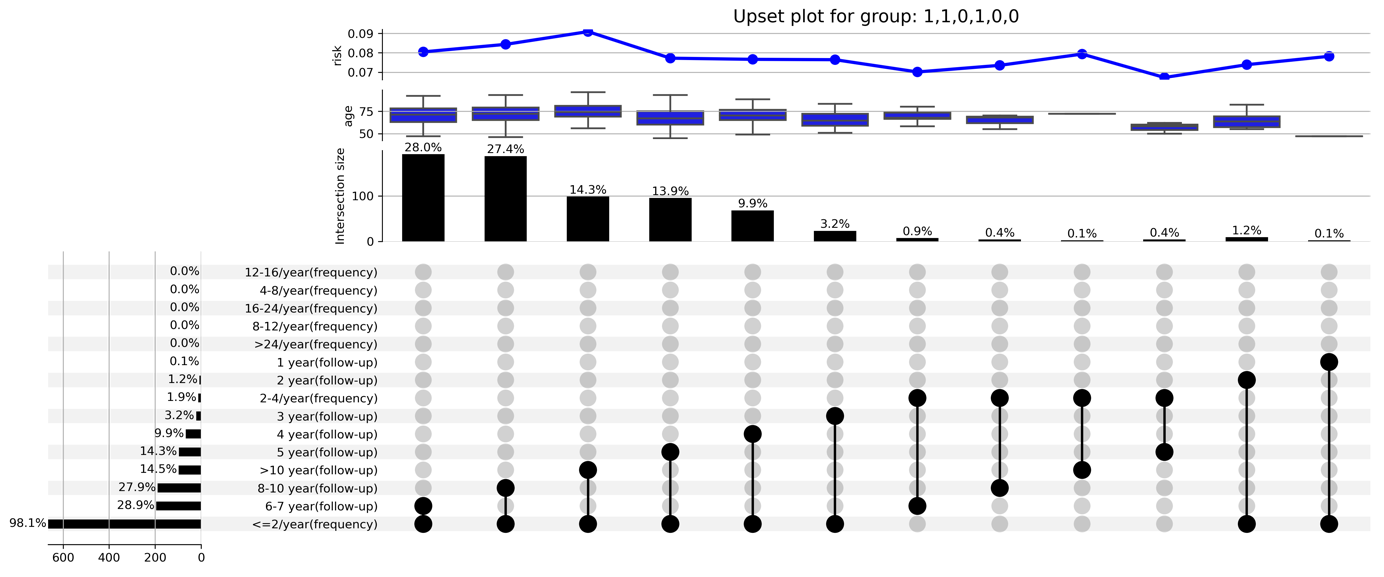}
 \centering
         \includegraphics[width=\textwidth]{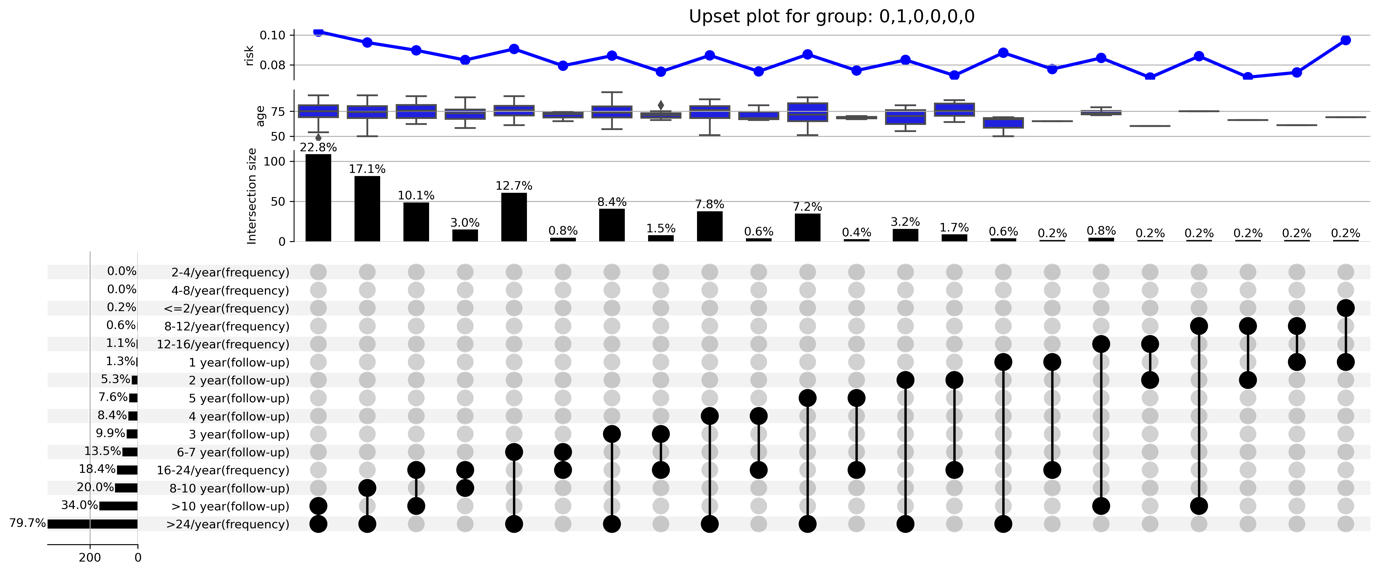}
\end{figure}
\begin{figure}[h!]
     \centering
         \includegraphics[width=\textwidth]{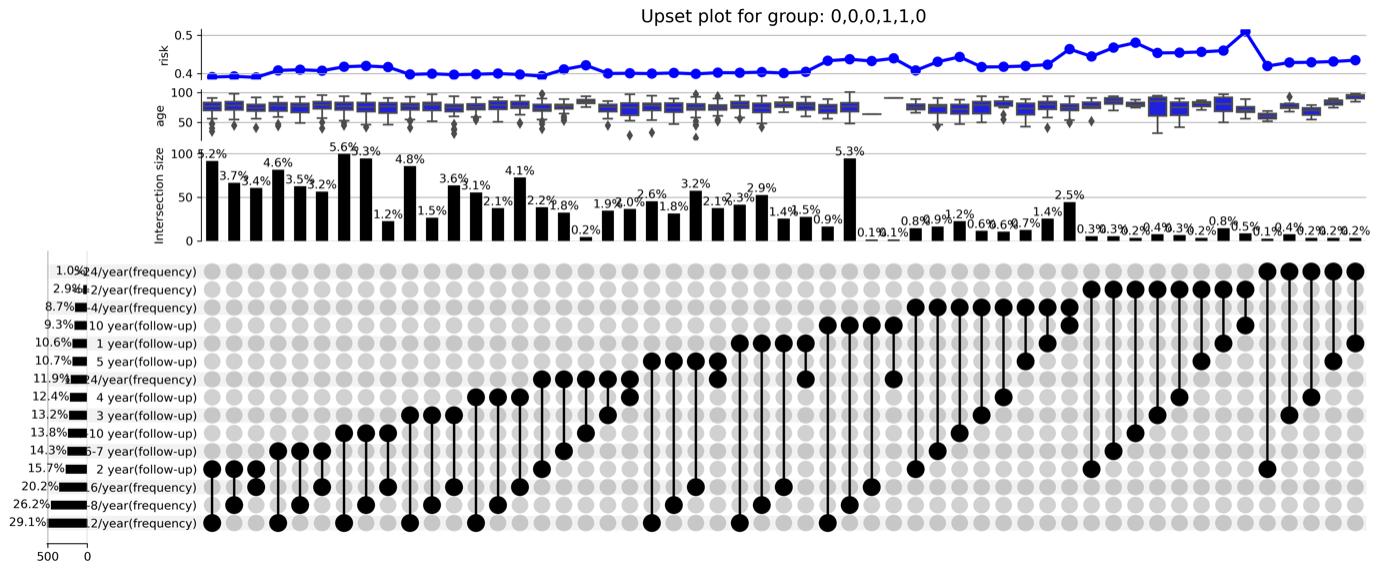}
         \caption{UpSet plots for F-HF}
\end{figure}

% \appendix
% \section{Do \emph{not} have an appendix here}

% \textbf{\emph{Do not put content after the references.}}
% %
% Put anything that you might normally include after the references in a separate
% supplementary file.

% We recommend that you build supplementary material in a separate document.
% If you must create one PDF and cut it up, please be careful to use a tool that
% doesn't alter the margins, and that doesn't aggressively rewrite the PDF file.
% pdftk usually works fine. 

% \textbf{Please do not use Apple's preview to cut off supplementary material.} In
% previous years it has altered margins, and created headaches at the camera-ready
% stage. 
%%%%%%%%%%%%%%%%%%%%%%%%%%%%%%%%%%%%%%%%%%%%%%%%%%%%%%%%%%%%%%%%%%%%%%%%%%%%%%%
%%%%%%%%%%%%%%%%%%%%%%%%%%%%%%%%%%%%%%%%%%%%%%%%%%%%%%%%%%%%%%%%%%%%%%%%%%%%%%%

\end{document}